%% file: main.tex
\documentclass[runningheads]{llncs}

\usepackage{eccv}

\usepackage{eccvabbrv}

\usepackage{graphicx}
\usepackage{booktabs}
\input{preamble}

\usepackage[accsupp]{axessibility}  

\usepackage{hyperref}

\usepackage{orcidlink}

\begin{document}

\title{Towards Consistent and Efficient Dataset Distillation via Diffusion-Driven Selection} 

\titlerunning{Dataset Distillation via Diffusion-Driven Selection}

\author{Xinhao Zhong\inst{1}\thanks{Equal Contribution} \and
{Shuoyang Sun\inst{1}}{*} \and
{Zhaoyang Xu\inst{1}}{*} \and
XUlin Gu\inst{2} \and \\
Bin Chen\inst{1}\thanks{Corresponding Author} \and
Min Zhang\inst{1} \and
Yaowei Wang\inst{1}\inst{3}}

\authorrunning{Xinhao Zhong et al.}

\institute{Harbin Institute of Technology, Shenzhen \and
Shanghai Jiao Tong University \and Pengcheng Laboratory}

\maketitle

\input{sec/0_abstract}
\input{sec/1_introduction}
\input{sec/2_related}

\input{sec/3_method}

\input{sec/5_experiment}
\input{sec/6_conclusion}

%
%
\bibliographystyle{splncs04}
\bibliography{main}

\clearpage
\input{sec/X_suppl}
\end{document}

%% file: preamble.tex
\usepackage[dvipsnames]{xcolor}

\usepackage{wrapfig}
\usepackage{multirow}
\usepackage{colortbl}

\usepackage{amsthm,amsmath,amssymb}
\usepackage{bbding}
\usepackage{algorithm}
\usepackage{algorithmic}
\usepackage{diagbox}

%% file: sec/0_abstract.tex
\begin{abstract}
Dataset distillation provides an effective approach to reduce memory and computational costs by optimizing a compact dataset that achieves performance comparable to the full original. However, for large-scale datasets and complex deep networks (e.g., ImageNet-1K with ResNet-101), the vast optimization space hinders distillation effectiveness, limiting practical applications. Recent methods leverage pre-trained diffusion models to directly generate informative images, thereby bypassing pixel-level optimization and achieving promising results. Nonetheless, these approaches often suffer from distribution shifts between the pre-trained diffusion prior and target datasets, as well as the need for multiple distillation steps under varying settings. To overcome these challenges, we propose a novel framework that is orthogonal to existing diffusion-based distillation techniques by utilizing the diffusion prior for patch selection rather than generation. Our method predicts noise from the diffusion model conditioned on input images and optional text prompts (with or without label information), and computes the associated loss for each image-patch pair. Based on the loss differences, we identify distinctive regions within the original images. Furthermore, we apply intra-class clustering and ranking on the selected patches to enforce diversity constraints. This streamlined pipeline enables a one-step distillation process. Extensive experiments demonstrate that our approach consistently outperforms state-of-the-art methods across various metrics and settings. The source code is available at \url{https://github.com/ndhg1213/DDDS}
\keywords{Dataset distillation \and Diffusion model}
\end{abstract}

%% file: sec/1_introduction.tex
\section{Introduction}
\label{sec:intro}

The rapid advancement of deep learning has driven impressive performance gains through increasingly deeper and more complex models trained on large-scale datasets~\cite{he2016deep,dosovitskiy2020image}. However, training these models demands significant memory and computational resources. Dataset distillation has recently emerged as an effective approach to data compression~\cite{lei2023comprehensive}, offering reductions in resource consumption. 

\begin{wrapfigure}{r}{0.55\linewidth}
    \centering
    \includegraphics[width=1\linewidth]{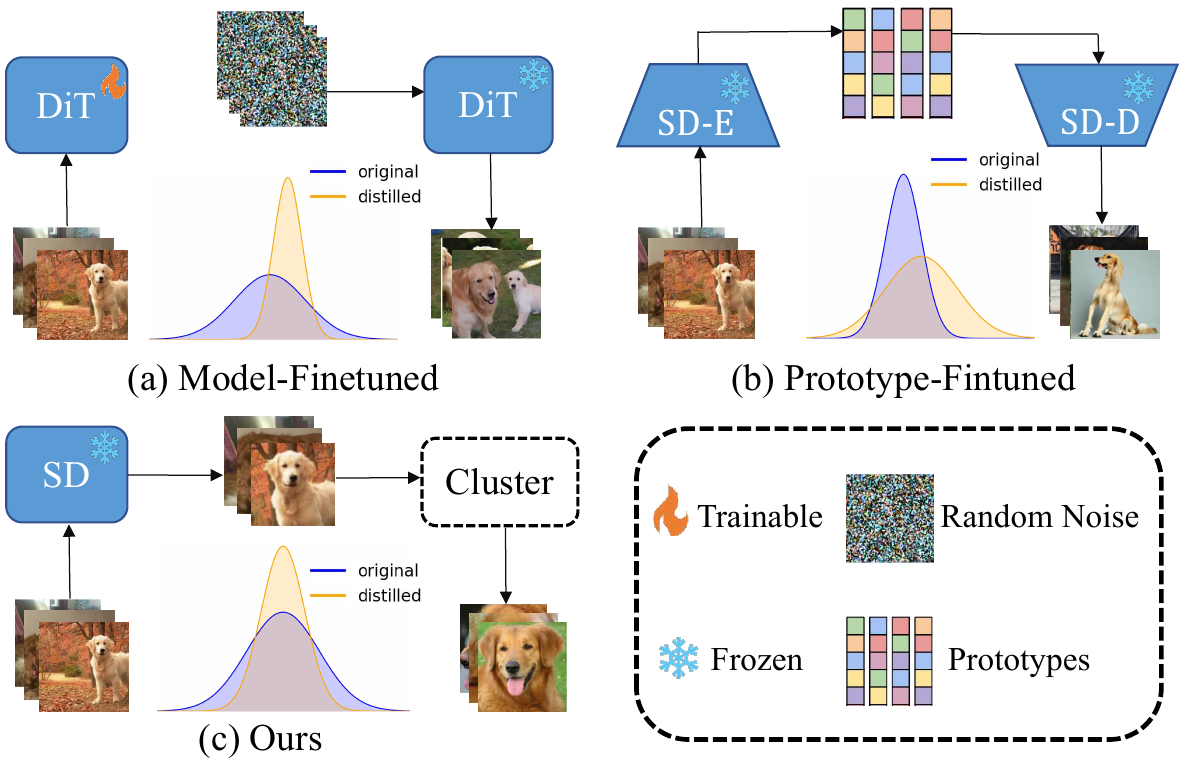}
    \caption{Comparison of diffusion-based distillation methods. (a) Fine-tuning a diffusion model pre-trained on ImageNet-1K to directly generate images. (b) Fine-tuning the prototypes of the original images to generate images. (c) Identifying the most representative regions in the real images. The images generated by our method more effectively represent the feature distribution of the original dataset.}
  \label{fig:intro}
\end{wrapfigure}
By optimizing images in pixel space through information matching (e.g., gradient matching), these optimization-based methods produce smaller synthetic datasets that can achieve comparable performance to the original data~\cite{zhao2021dsa,zhao2021datasetdm,cazenavette2022dataset,zhong2025going,gu2025temporal}. Despite their promising performance, the vast pixel-level optimization space restricts the feasibility of these methods to small-scale datasets, e.g.,  CIFAR-10/100~\cite{krizhevsky2009learning}, thus limits their generalization.
A recent approach, SRe$^2$L~\cite{yin2024squeeze}, scales optimization-based methods to larger datasets (e.g., ImageNet-1K~\cite{deng2009imagenet}) by using a pre-trained classifier to guide dataset synthesis. 
However, its strong dependence on proxy models during optimization often hinders the distilled dataset's generalization, limiting broader applicability. Another critical limitation arises when handling various distillation configurations, such as different subsets of large datasets. Previous optimization-based methods typically incur substantial redundant costs due to repeated distillation requirements. Although recent works~\cite{he2024multisize,he2024you} aim to address this issue, their applicability remains confined to small-scale datasets.

Subsequent methods~\cite{sun2024diversity,gu2024efficient,su2024d,chen2025influence,chan2025mgd,zhao2025taming} that avoid direct image optimization have shown promise for scaling dataset distillation to larger datasets and more complex model architectures, addressing several severe limitations of previous methods. RDED~\cite{sun2024diversity} leverages a pre-trained classifier (e.g., ResNet-18~\cite{he2016deep}) to select patches randomly cropped from real images and then concatenate them into synthetic images. While RDED achieves substantial performance improvements, it requires careful model selection and retraining of a classifier tailored to the specific dataset. Moreover, concatenated images often lack complete class-relevant features, especially for complex models with higher learning capacity. Model-Finetuned methods~\cite{gu2024efficient,chen2025influence} address distillation by pruning the DiT model~\cite{peebles2023scalable} to generate synthetic datasets using regularization based on dataset representativeness and diversity. However, these methods rely on a diffusion model pre-trained on ImageNet-1K, requiring fine-tuning for different target datasets, which limits flexibility. Prototype-Finetuned methods~\cite{su2024d,chan2025mgd} apply a pre-trained text-to-image Stable Diffusion (SD)\cite{rombach2022high} to generate synthetic datasets by finetuning original images in latent space. However, without constraints on distributional differences between the LDM’s pre-training data (e.g., LAION\cite{schuhmann2022laion}) and the target dataset (e.g., ImageNet-1K), these methods can produce class-irrelevant or noisy images.

To overcome these limitations, we propose a novel dataset distillation paradigm that 
leverages diffusion model’s prior to select the most informative image patches. Specifically, for each input image from the original dataset, we first predict the noise map using the pre-trained SD, with and without the label text prompt, calculating losses for each scenario. The differential loss quantifies representativeness, enabling us to crop patches most relevant to the label guided by the target dataset’s data distribution.
Additionally, we can effectively constrain the implicit data distribution learned by the diffusion model by using the pre-trained SD to identify class-pertinent regions.

To mitigate the inherent diversity of diffusion models, we apply clustering to capture the visual features most representative of each class. Specifically, we compute diffusion features for all patches inspired by recent work~\cite{tang2023emergent}, allowing for efficient clustering and ranking of the cropped patches. \Cref{fig:intro} illustrates a comparison between our method and previous diffusion-based generative distillation methods. Extensive experiments demonstrate that our approach achieves state-of-the-art (SOTA) performance on large-scale datasets, supporting an efficient, unrestricted, one-step distillation process.

Our contributions are summarized as follows:
\begin{itemize}
\item We introduce a novel dataset distillation framework that diverges from conventional methods by leveraging pre-trained diffusion models in an orthogonal way. Our approach uniquely addresses distributional discrepancies between the diffusion model and target dataset, offering a new perspective on improving dataset fidelity and applicability. 
\item Our proposed paradigm enables a highly efficient, one-step distillation process, eliminating the need for model training and fine-tuning, and avoiding repetitive distillation across class combinations or compression ratios. 
\item Extensive experiments demonstrate that our method consistently surpasses existing training-free methods across various settings.
\end{itemize}

%% file: sec/2_related.tex
\section{Related Work}

\subsection{Dataset Distillation}

Dataset distillation was initially regarded as a meta-learning problem~\cite{wang2018dataset,li2025dd,zhong2025rectified}. It involves minimizing the loss function of the synthetic dataset through a model trained on this dataset. Since then, several approaches have been proposed to enhance the performance of dataset distillation. 


Recent methods propose the decoupled distillation strategy and successfully extend the dataset distillation application to large-scale datasets. SRe$^2$L~\cite{yin2024squeeze} leverages a pre-trained classifier to compute the cross-entropy loss of the synthetic dataset for updates, while RDED~\cite{sun2024diversity} calculates the confidence of randomly cropped patches from the original images with a pre-trained classifier and concatenates them together to form the synthetic dataset. 
Recent methods~\cite{cazenavette2023generalizing,zhong2024hierarchical} have introduced GAN as a parameterization technique. 
While with the development of generative models, the diffusion models are introduced to directly generate synthetic datasets, achieving promising performance~\cite{gu2024efficient,su2024d,chen2025influence,zhao2025taming,chan2025mgd}. However, existing diffusion-based methods are constrained by the implicit data distribution learned by diffusion models, leading to significant distributional bias in the synthetic dataset. In addition, aforementioned methods still fail to complete the distillation process in one pass for all settings. To address these issues, we utilize diffusion model to actively identify the most class-relevant regions in the original images, enabling an efficient dataset distillation process.

\subsection{Diffusion Model}

As a class of generative approaches, diffusion-based frameworks learn to progressively convert Gaussian-distributed noise vectors $\epsilon \sim \mathcal{N}(\mathbf{0}, \mathbf{I})$ into samples matching target distributions $\mathcal{X}$, demonstrating remarkable effectiveness across diverse applications~\cite{zhu2024multibooth,zhu2024instantswap,siglidis2024diffusion}. These models employ a neural network $\epsilon_\theta(\mathbf{x},t)$ parameterized by $\theta$ to estimate noise components, operating through sequential refinement steps denoted by a temporal parameter $t$. The denoising procedure typically involves iterative updates where the network processes corrupted inputs $\mathbf{x}$ at progressive noise levels determined by different timestap $t$.

The training protocol for latent diffusion models involves two core phases. Initially, an encoder network $E(\cdot)$ compresses input samples $\mathbf{x}$ into latent representations $\mathbf{z} = E(\mathbf{x})$. These latent codes undergo systematic corruption through a diffusion trajectory determined by uniformly sampled timestep $t$, implemented via the forward process formulation:
\begin{align}
N_{t}(\mathbf{z}, \epsilon, t) = \sqrt{\bar{a}{t}}\mathbf{z} + (1-\sqrt{\bar{a}{t}})\epsilon,
\end{align}
where $N_{t}$ denotes the predicted noise at timesteps $t$, and the time-dependent attenuation factors $\sqrt{\bar{a}_t}$ govern the noise blending proportions.
The denoising network $\epsilon_\theta$ then learns to recover original signals, which is achieved through the optimization objective:
\begin{align}
\label{eq:ldm_loss}
\mathcal{L}_t(\mathbf{z}, \epsilon) = |\epsilon_\theta(N_{t}(\mathbf{z}, \epsilon, t), t) - \epsilon|_2^2.
\end{align}

During the generation phase, novel samples are synthesized through a reverse diffusion process~\cite{song2020denoising,ho2020denoising} that progressively refines Gaussian noise samples $\mathbf{z}_T$ over multiple iterations. For conditional generation tasks like text-to-image synthesis, the framework incorporates semantic conditioning signals $c$ (e.g., text embeddings) as auxiliary inputs. This extends the denoiser architecture to $\epsilon_\theta(\mathbf{z},t,c)$, modifying the training objective accordingly:
\begin{align}
\mathcal{L}_t(\mathbf{z}, \epsilon, c) = |\epsilon_\theta(N_{t}(\mathbf{z}, \epsilon, t), t, c) - \epsilon|_2^2.
\end{align}

%% file: sec/3_method.tex
\begin{figure*}[t]
  \centering
  \includegraphics[width=0.95\linewidth]{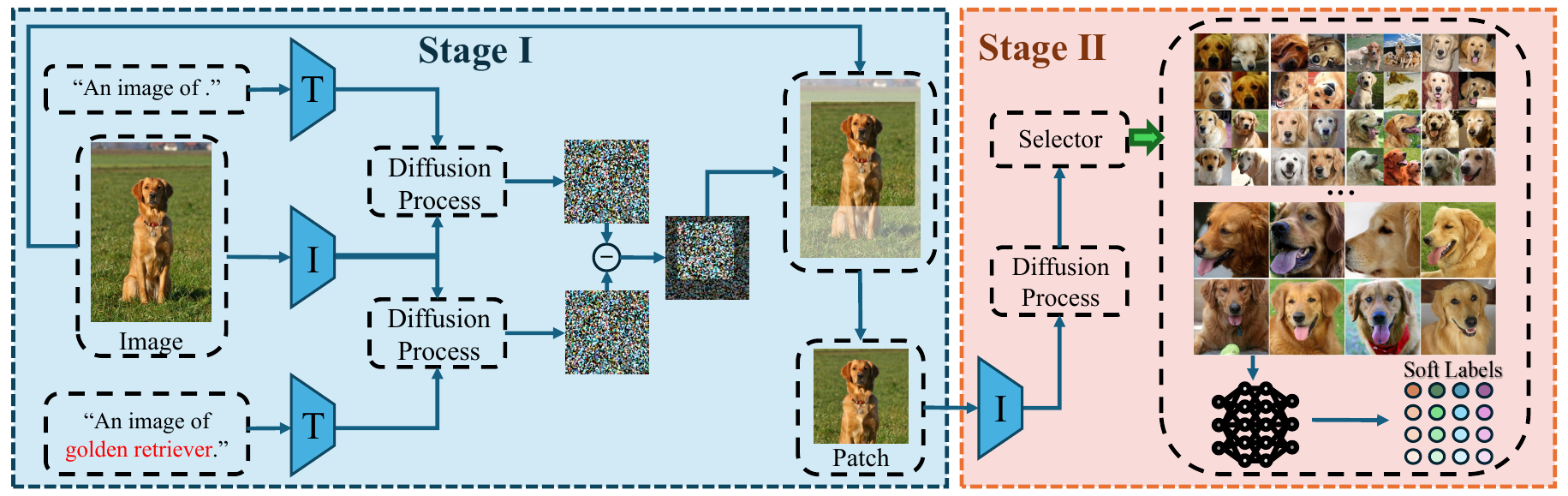}
  \caption{(a): Stage $\textup{\uppercase\expandafter{\romannumeral1}}$ leverages image and corresponding differential class text prompts as inputs to the diffusion model to select the regions that most represent class-relevant features. (b): Stage $\textup{\uppercase\expandafter{\romannumeral2}}$ first computes DIFT of the patches and performs clustering to aggregate the most representative visual elements for each class. The evaluation process utilizes soft labels provided by a pre-trained teacher model. Rather than existing methods employing diffusion models, our proposed method maximizes the restoration of the original data distribution.}
  \label{fig:framework}
  \vspace{-0.8em}
\end{figure*}
\section{Method}

In this section, we delve into the specifics of our method, aiming to address the limitations of previous methods. The overall framework of our method is shown in \Cref{fig:framework}.

\subsection{Optimal Class-relevant Feature Selection}

In recent years, several methods propose the utilization of pre-trained diffusion models to directly generate images in order to improve dataset distillation performance on large-scale datasets. However, these methods often face the challenge of distributional differences between the pre-training dataset and the target dataset. Model-finetuned methods~\cite{gu2024efficient,chen2025influence} rely on fine-tuning a diffusion model pre-trained on ImageNet-1K, while Prototypes-finetuned methods~\cite{su2024d,chan2025mgd} attempt to obtain representative samples through finetuning the original images in latent space. Both of them remain constrained by the distributional shifts introduced during the unconstrained denoising process.

To address the issue of data heterogeneity, we introduce a framework consisting of two stages that utilizing the diffusion model for localization rather than generation. As illustrated in Stage $\textup{\uppercase\expandafter{\romannumeral1}}$ of \Cref{fig:framework}, we first utilize diffusion models as zero-shot image-wise classifiers~\cite{li2023your}, which can be formulated as:
\begin{equation}
\begin{aligned}
\label{eq:multi}
p(c_{i} | \mathbf{z}) & = \frac{\mathrm{exp}\{-\mathbb{E}_{\epsilon, t}[
\epsilon_\theta(N_{t}(\mathbf{z}, \epsilon, t),\; t,\; c_{i}) - \epsilon]\}}{\sum_{j}\mathrm{exp}\{-\mathbb{E}_{\epsilon, t}[
\epsilon_\theta(N_{t}(\mathbf{z}, \epsilon, t),\; t,\; c_{j}) - \epsilon]\}} \\
& = \frac{1}{\sum_{j}\mathrm{exp}\{ \mathbb{E}_{\epsilon, t}[
N_t(\mathbf{z}, \epsilon, c_{i}) - N_t(\mathbf{z}, \epsilon, c_{j})] \}}.\\
\end{aligned}
\end{equation}
where $c_{i} \in [C]$ and $[C]$ represents the label space, $\mathbf{z}$ is the latent of input image $\mathbf{x}$. However, such an approach can only select the image with the highest classification probability, lacking further constraints. Additionally, a larger class space can lead to significant computational overhead. Using the classifier-free guidance\cite{ho2022classifier}, we redefine the label space for a specific label $c$ and rewrite \Cref{eq:multi} as follow:
\begin{align}
\label{eq:single}
p(c | \mathbf{z}) = \frac{1}{1 + \mathrm{exp}\{ \mathbb{E}_{\epsilon, t}[
N_t(\mathbf{z}, \epsilon, c) - N_t(\mathbf{z}, \epsilon, \phi)] \}},
\end{align}
where $\phi$ and $c$ represents the classifier-free guidance and label text prompt respectively (i.e., ``An image of .'' and ``An image of c''). We then simplify the computation as follow:
\begin{align}
\label{eq:repre}
M(\mathbf{z} | c) = \mathbb{E}_{\epsilon, t}[
N_t(\mathbf{z}, \epsilon, c) - N_t(\mathbf{z}, \epsilon, \phi)].
\end{align}
Obviously \Cref{eq:single} and \Cref{eq:repre} are monotonically equivalent. Based on the differential noise map $M(\mathbf{z} | c) \in \mathbb{R}^{H\times W\times3}$, we could calculate the representative score $R(\mathbf{p} | c)$ within each patch $\mathbf{p}$ in original image.
To enable computation on the original image, we first upsample the noise map from the latent space to the original image resolution via interpolation and take the channel-wise average to obtain $\hat{M}(\mathbf{x} | c) \in \mathbb{R}^{H\times W}$. We then apply an average pooling kernel to compute the score for each patch, as illustrated below:

\begin{align}
\label{eq:patch}
R(\mathbf{p} | c) = \sum_{\mathbf{x}_{i}, \mathbf{x}_{j} \in \mathbf{p}}\hat{M}(\mathbf{x} | c)[i, j].
\end{align}

Based on these representation scores, we can select the final distilled data and further reduce the complexity. The detailed workflow is depicted in \Cref{alg:our}. 

\begin{algorithm}[!t]
\caption{Pseudocode of the proposed method}
\begin{algorithmic}[1]
    \REQUIRE ($\mathcal{T}$, $\mathcal{Y}$): Real dataset and corresponding label texts. $\mathcal{P}$: Selected patches. $I$: Pre-trained image encoder. $T$: Pre-trained text encoder. $\mathcal{U}$: Pre-trained time-conditional U-Net. $T$: Number of diffusion time-steps. $C$: Number of top cluster centers.\\
    \STATE initialize $\mathcal{S} = \emptyset$
    \STATE $Z=I(\mathcal{T})\sim P_z$
    \FOR{ each class $Y\in\mathcal{Y}$}
    \STATE $c=T(Y)$
    \STATE initialize $\mathcal{S}_{Y}, \mathcal{P}_{Y} = \emptyset$
    \FOR{ each latent $\mathbf{z}\in Z$}
    \FOR { i ${\leftarrow}$ 0 to $N - 1$}
    \STATE $t \sim U(0.1, 0.7)$
    \STATE $N_{i} = N_{t}(\mathbf{z}, \epsilon, c) - N_{t}(\mathbf{z}, \epsilon, \phi)$
    \ENDFOR
    \STATE $M(\mathbf{z}|c) = \frac{1}{T} \sum_{i} N_{i}$
    \STATE $\mathbf{p} = \mathop{\arg\max}\limits_{\mathbf{p} }\sum_{\mathbf{x}_{i}, \mathbf{x}_{j} \in \mathbf{p}}\hat{M}(\mathbf{x} | c)[i, j])$
    \STATE $\mathcal{P}_{Y} \leftarrow  \mathcal{P}_{Y} \cup \{\mathbf{p}\}$
    \ENDFOR
    \STATE $cluster(\mathcal{U}(I(\mathcal{P}_{Y}), c, 160))$
    \FOR{i ${\leftarrow}$ 0 to $C - 1$}
    \STATE $\mathcal{S}_{Y} \leftarrow \mathcal{S}_{Y} \cup top(\mathcal{P}_{Y}^{i})$
    \ENDFOR
    \STATE $\mathcal{S} \leftarrow \mathcal{S} \cup \mathcal{S}_{Y}$\\
    \ENDFOR
    \ENSURE $\mathcal{S}$: Distilled dataset.
\end{algorithmic}
\label{alg:our}
\end{algorithm}

\subsection{Visual Elements Aggregation}

Although we have successfully selected the most representative patches, the overly complex data often introduce excessive randomness. To enhance the representativeness and reduce the randomness of the synthetic dataset, we perform an effective selecting strategy on the cropped patches as shown in the Stage $\textup{\uppercase\expandafter{\romannumeral2}}$ of \Cref{fig:framework}. With diffusion model, we select DIFT~\cite{tang2023emergent}for applying clustering, where noise at a specific time step is added to the latent of input image, and passed through a U-Net~\cite{ronneberger2015u} with text condition to obtain the intermediate layer activations as features.
On this basis, we use K-means~\cite{lloyd1982least} to cluster the selected patches. 

To better leverage clustering information, we have designed a ranking rule: \textit{\textbf{1) Intra-cluster.}} patches are ranked based on their distance to the centroid instead of corresponding $R(\mathbf{p}|c)$. \textit{\textbf{2) Inter-cluster.}} clusters are ranked based on the median $R(\mathbf{p}|c)$ of the patches within each cluster. 
In practice, we fix the number of cluster centers at 32 and prioritize the selection of patches from the top 10 clusters under all compression settings to ensure both representativeness and diversity. 
Furthermore, when IPC $\leq 10$, we follow the experimental setup of RDED by concatenating multiple patches into a single image. While for other settings, we treat each patch as an individual image.

%% file: sec/5_experiment.tex
\begin{figure*}[!t]
    \centering
    \resizebox{\textwidth}{!}{
    \includegraphics{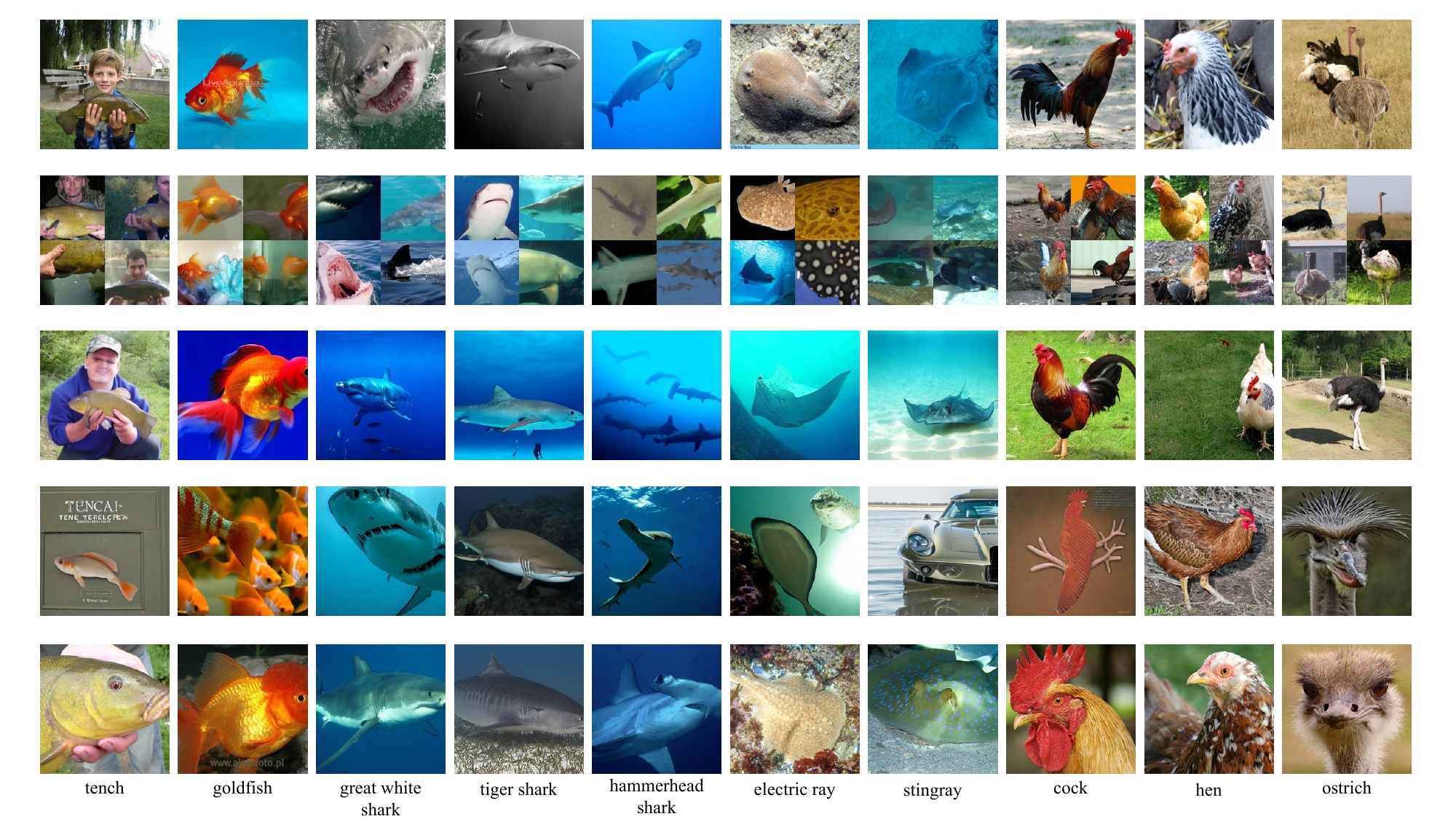}}
    \caption{Visualization comparison of images from the top ten classes of the ImageNet-1k dataset is presented (not cherry picked). From top to bottom, each row corresponds to the real dataset, RDED, Minimax, D$^4$M, and ours under IPC=50. Compared to existing methods, our method fully adheres to the distribution of the real dataset and demonstrates a sharper focus on class-specific features. }
    \label{visualization}
\end{figure*}

\section{Experiments}

In this section, we demonstrate that our method outperforms existing approaches in both performance and computational efficiency. More experimental results and additional ablation studies are provided in the appendix.

\subsection{Experimental Settings}

\textbf{Datasets and Architectures.}
To evaluate our proposed method on low-resolution datasets, we  conduct experiments on CIFAR-10/100~\cite{krizhevsky2009learning} and TinyImageNet~\cite{le2015tiny}, with ConvNet~\cite{gidaris2018dynamic} serving as the backbone. For high-resolution data, we perform evaluations on ImageNet-1K~\cite{deng2009imagenet} and its widely used subsets, including ImageNet-100~\cite{kim2022dataset}, ImageNette and ImageWoof~\cite{cazenavette2022dataset}, utilizing ResNet-18~\cite{he2016deep} as the backbone.

\textbf{Baselines.} we consider recent methods applicable to large-scale datasets as strong competitors.  SRe$^2$L~\cite{yin2024squeeze}, by updating the synthetic dataset using the CE-loss of a pre-trained classifier and introducing soft label matching, was the first method to be feasibly applied to large-scale datasets. RDED~\cite{sun2024diversity} significantly improved computational efficiency and performance by evaluating randomly cropped patches with a pre-trained classifier and concatenating them together to generate images. Minimax~\cite{gu2024efficient} and IGD~\cite{chen2025influence} proposed fine-tuning DiT model pre-trained on ImageNet-1K by using the distribution of target dataset as a regularization to directly generate images. D$^4$M~\cite{su2024d} and MGD$^3$~\cite{chan2025mgd} extended the range of diffusion models to pre-trained text-to-image Stable Diffusion, utilizing the cluster centers of latent to generate synthetic datasets.

\textbf{Evaluation Protocol.}
In line with previous work, we set IPC=10 and 50 for low-resolution datasets, and IPC=10, 50, and 100 for high-resolution datasets to enable more efficient learning. During the training process, we employ a fixed pre-trained classifiers to provide soft labels. Subsequently, five randomly initialized networks are trained on the distilled dataset, with the average performance of these networks evaluated. We use the same evaluation hyper-parameters with EDC~\cite{shao2024elucidating}.

\textbf{Implementation Details.}
When calculating $M(\mathbf{z}|c)$ of images, we select a time-step range of [0.1, 0.7], whereas for the computation of DIFT, a time-step of 160 is employed. Across all IPC settings, we prioritize cropping the patches with resolution of $224 \times 224$ and selecting them from the top 10 ranked clusters after clustering. All the experiments were carried out using PyTorch on NVIDIA RTX-3090 GPUs. 

\begin{table*}[!ht]
    \centering
    \caption{Performance comparison with methods applicable to large-scale datasets. Following the evaluation protocol of RDED, we use ResNet-18 as the teacher model for generating the dataset.}
    \scalebox{0.8}{
        \begin{tabular}{@{}cc|cccccccc}
            \toprule
            &     & \multicolumn{2}{c}{ResNet-18} & \multicolumn{3}{c}{DiT-ImageNet} & \multicolumn{3}{c}{Stable Diffusion}\\ 
            \cmidrule(lr){3-4} \cmidrule(lr){5-7} \cmidrule(lr){8-10}
            Dataset       & IPC  & SRe$^2$L  & RDED   & Minimax                    &IGD   & D$^{3}$HR    & D$^4$M     &  MGD$^{3}$    & Ours  \\  \midrule
                          & 10   & 29.4 $\pm$ 3.0   & 61.4 $\pm$ 0.4   & 57.2 $\pm$ 0.8  &  58.3 $\pm$ 0.2 & 58.1 $\pm$ 0.2  & 57.4 $\pm$ 0.4  & 58.9 $\pm$ 0.4 & \textbf{61.7 $\pm$ 0.4}  \\
            ImageNette    & 50  & 40.9 $\pm$ 0.3  & 80.4 $\pm$ 0.4   &  84.4 $\pm$ 0.5   & 85.5 $\pm$ 0.3  & 85.7 $\pm$ 0.2  & 84.8 $\pm$ 0.2   & 85.2 $\pm$ 0.5 & \textbf{86.4 $\pm$ 0.2}  \\
                          & 100  &  50.2 $\pm$ 0.4 &  89.6 $\pm$ 1.0 &  90.1 $\pm$ 0.6 &   91.6 $\pm$ 0.5  &  90.9 $\pm$ 0.2 & 90.4 $\pm$ 0.7  & 91.0 $\pm$ 0.2 & \textbf{92.0 $\pm$ 0.3}  \\ \midrule
                          & 10   & 20.2 $\pm$ 0.2 & 38.5 $\pm$ 2.1 & 38.1 $\pm$ 0.6 & 39.8 $\pm$ 0.1 & 39.3 $\pm$ 0.3 & 36.8 $\pm$ 1.2  & 38.2 $\pm$ 0.4 & \textbf{44.5 $\pm$ 0.4}   \\ 
            ImageWoof     & 50  & 23.3 $\pm$ 0.3 & 68.5 $\pm$ 0.7 & 71.2 $\pm$ 0.6 & 72.8 $\pm$ 0.4 & 73.0 $\pm$ 0.2 & 71.4 $\pm$ 1.4  & 72.1 $\pm$ 0.5 & \textbf{73.5 $\pm$ 1.4}      \\
                          & 100  & 37.6 $\pm$ 0.3 & 77.2 $\pm$ 0.6 &  77.4 $\pm$ 0.2 & 78.3 $\pm$ 0.4 & 78.7 $\pm$ 0.3 & 78.5 $\pm$ 0.3  & 77.9 $\pm$ 0.2 & \textbf{79.6 $\pm$ 0.2} \\ \midrule
                          & 10   & 9.5 $\pm$ 0.4 & 36.0 $\pm$ 0.3  & 31.6 $\pm$ 0.7 & 33.2 $\pm$ 0.3 & 34.0 $\pm$ 0.2  & 33.4 $\pm$ 0.2 & 34.2 $\pm$ 0.5  & \textbf{36.8 $\pm$ 0.2}   \\
            ImageNet-100  & 50  & 27.0 $\pm$ 0.4 & 61.6 $\pm$ 0.1 & 64.0 $\pm$ 0.5 & 64.2 $\pm$ 0.3 & 65.3 $\pm$ 0.2 & 62.7 $\pm$ 0.6   & 64.1 $\pm$ 0.3 & \textbf{67.5 $\pm$ 0.3 }\\
                          & 100  & 39.4 $\pm$ 0.1 & 75.2 $\pm$ 0.3 & 76.4 $\pm$ 0.3 & 77.8 $\pm$ 0.3 & 77.2 $\pm$ 0.2 & 76.8 $\pm$ 0.5  & 76.7 $\pm$ 0.3 & \textbf{78.6 $\pm$ 0.2 }  \\ \midrule
                          & 10   & 21.3 $\pm$ 0.6 & 42.0 $\pm$ 0.1 & 44.3 $\pm$ 0.3 & 45.5 $\pm$ 0.5 & 44.3 $\pm$ 0.3 & 27.9 $\pm$ 0.3  & 41.8 $\pm$ 0.2  & \textbf{46.1 $\pm$ 0.3} \\
            ImageNet-1K   & 50  & 46.8 $\pm$ 0.2 & 56.5 $\pm$ 0.1 & 58.6 $\pm$ 0.3 & 60.3 $\pm$ 0.4 & 59.4 $\pm$ 0.1 & 55.2 $\pm$ 0.3  & 57.2 $\pm$ 0.4 & \textbf{61.0 $\pm$ 0.3}  \\
                          & 100  & 52.5 $\pm$ 0.5 & 59.5 $\pm$ 0.8 & 60.2 $\pm$ 0.3 & 61.4 $\pm$ 0.3 & 62.5 $\pm$ 0.0 & 59.3 $\pm$ 0.4  & 61.7 $\pm$ 0.5 & \textbf{63.0 $\pm$ 0.2}     \\
                          \bottomrule
        \end{tabular}
    }
    \label{tab:high}
\end{table*}
\begin{table*}[t]
    \centering
    \caption{Performance comparison on low-resolution datasets. To compare with information matching distillation methods, we use ConvNet-3 and ConvNet-4 as the backbone for CIFAR10/100 and TinyImagenet respectively. For decoupled distillation methods, we use ResNet-18 as the backbone. All the performance are evaluated on the same model architecture.
    }
    \scalebox{0.77}{
        \begin{tabular}{@{}cc|ccccc|ccc@{}}
            \toprule
                          &     & \multicolumn{5}{c|}{ConvNet} & \multicolumn{3}{c}{ResNet-18}                                                                                                                                                   \\ \cmidrule(lr){3-7} \cmidrule(lr){8-10} 
            Dataset       & IPC & DSA                          & IDM                           & TESLA                            &RDED           & Ours          & SRe$^2$L   & D$^4$M    &  Ours \\  \midrule
            \multirow{2}{*}{CIFAR10}      & 10  & 52.1 $\pm$ 0.5               & 58.6 $\pm$ 0.1                & 66.4 $\pm$ 0.8                  & \textbf{50.2 $\pm$ 0.3}          & 48.8 $\pm$ 0.4 & 29.3 $\pm$ 0.5 & 34.2 $\pm$ 0.3 & \textbf{36.4 $\pm$ 0.2}\\
                          & 50  & 60.6 $\pm$ 0.5               & 67.5 $\pm$ 0.1                & 72.6 $\pm$ 0.7                 & \textbf{68.4 $\pm$ 0.1}  & 66.2 $\pm$ 0.3         & 45.0 $\pm$ 0.7 & 60.9 $\pm$ 0.1 & \textbf{61.0 $\pm$ 0.6}\\ \midrule
            \multirow{2}{*}{CIFAR100}     & 10  & 32.3 $\pm$ 0.3               & 45.1 $\pm$ 0.1                & 41.7 $\pm$ 0.3                       & \textbf{48.1 $\pm$ 0.3} & 47.3 $\pm$ 0.4 & 27.0 $\pm$ 0.4 & 40.8 $\pm$ 0.5 & \textbf{41.5 $\pm$ 0.4} \\
                          & 50  & 42.8 $\pm$ 0.4               & 50.0 $\pm$ 0.2                & 47.9 $\pm$ 0.3                     & 57.0 $\pm$ 0.1 & \textbf{57.6 $\pm$ 0.2}  & 50.2 $\pm$ 0.4 & 62.3 $\pm$ 0.4 & \textbf{63.8 $\pm$ 0.3} \\ \midrule
            \multirow{2}{*}{TinyImageNet} & 10  & -        & 21.9 $\pm$ 0.3                & -                                & 39.6 $\pm$ 0.1 & \textbf{40.3 $\pm$ 0.2} & 16.1 $\pm$ 0.2 & 35.1 $\pm$ 0.4 & \textbf{40.2 $\pm$ 0.5}\\
                          & 50  & -         & 27.7 $\pm$ 0.3                & -             & 47.6 $\pm$ 0.2 & \textbf{49.2 $\pm$ 0.7} & 41.1 $\pm$ 0.4 & 52.7 $\pm$ 0.4 & \textbf{58.5 $\pm$ 0.2}  \\ 
                          \bottomrule
        \end{tabular}
    }
    \label{tab:low}
\end{table*}

\subsection{Performance Improvements}

\textbf{High-resolution Results.}
For high-resolution datasets, we first resize the smallest edge of images to 256 pixels while maintaining height-width ratio. The resized images are then fed into the diffusion model to calculate $M(\mathbf{z}|c)$ of the images and that of all patches with a $224 \times 224$ averaging pool kernel. 
For instance, when IPC = 100, we select 10 patches from the top ten clusters. If a cluster contains insufficient patches, we supplement the synthetic dataset by selecting the highest-scoring patches. As shown in \Cref{tab:high}, our proposed method demonstrates significant improvements across various datasets and IPC settings. In specific configurations, our approach exceeds baseline's performance by more than 10\%, further validating the superiority of our method.

\textbf{Low-resolution Results.}
For low-resolution datasets, we maintain the same setting as in RDED, utilizing the diffusion model to calculate the $M(\mathbf{z}|c)$ of full-image without cropping, followed by clustering, we then evaluate the synthetic datasets on the same architecture. As demonstrated in \Cref{tab:low}, the incorporation of the diffusion model yields performance on CIFAR-10 that is comparable to, or even exceeds, that of existing methods when applied within our proposed framework. On more complex datasets, such as CIFAR-100 and TinyImageNet, our approach outperforms all current optimization-based methods and achieves SOTA performance, surpassing all prior techniques.

\begin{figure}[!t]
  \centering
  \includegraphics[width=\linewidth]{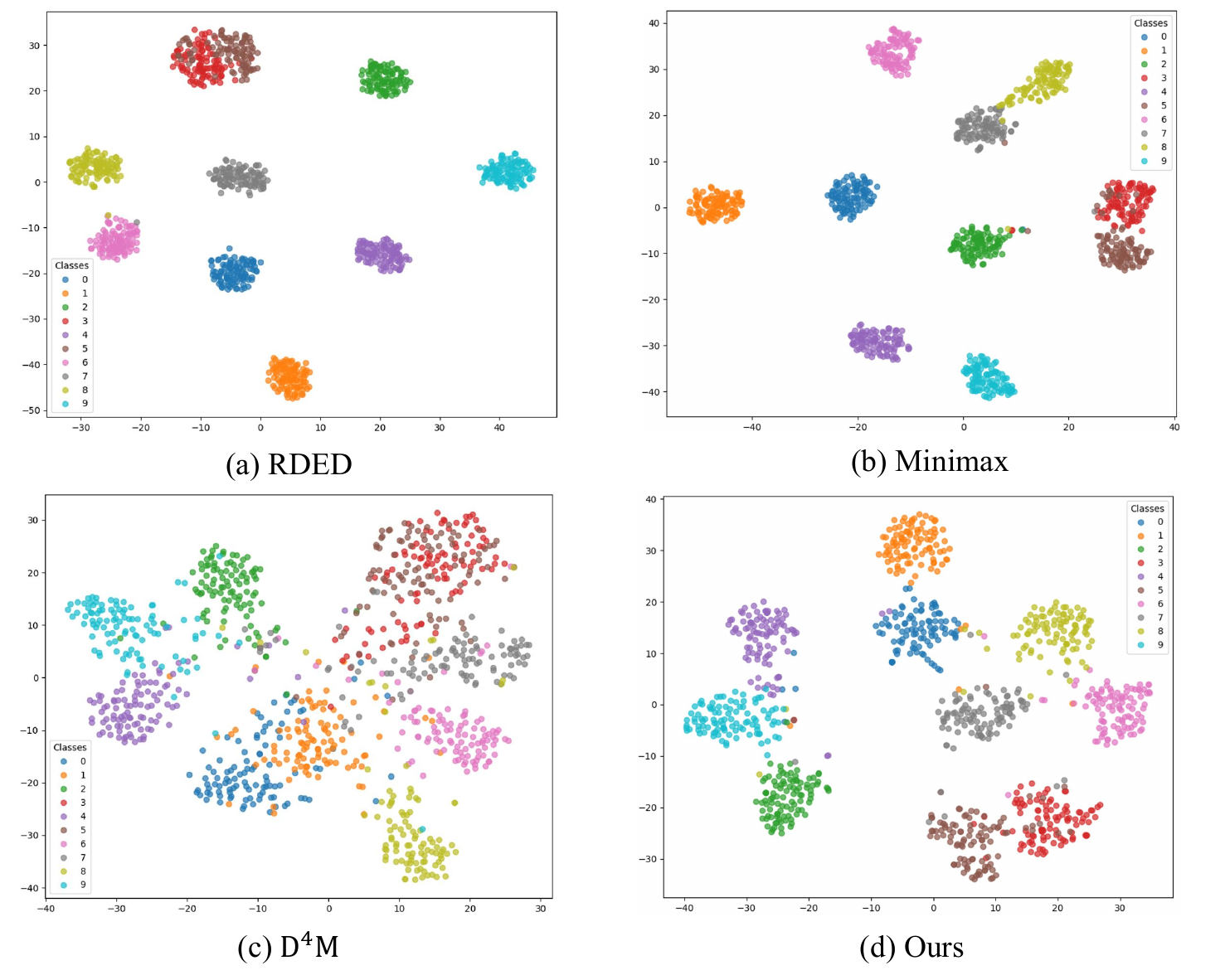}
  \caption{Comparison of t-SNE visualizations of features extracted by a pre-trained ResNet-18 on ImageWoof under IPC=100.}
  \label{fig:vis-tsne}
\end{figure}
\subsection{Qualitative Analysis}
To better understand the superiority of the datasets generated by our proposed method, we used t-SNE to visualize and analyze datasets produced by different methods. As shown in \Cref{fig:vis-tsne}, it is evident that the dataset generated by RDED \cite{sun2024diversity} is overly concentrated around class centers because it originates from a pre-trained classifier. 
Additionally, since the pre-trained classifier cannot guarantee accurate classification results, the synthetic dataset from RDED suffers from class mixing issues. 
Similarly, the Minimax \cite{gu2024efficient} and IGD \cite{chen2025influence}, which uses the DiT pre-trained on ImageNet-1K to generate images, also results in datasets that are overly concentrated in class centers. 
This concentration can be detrimental to model learning when IPC increases. 
On the other hand, D$^4$M \cite{su2024d} and MGD$^3$ \cite{chan2025mgd} generates images using SD, which is entirely unrelated to ImageNet-1K, resulting in datasets with excessive noise and outlier samples. 
In contrast to all the aforementioned methods, our approach achieves a good balance between diversity and representativeness by using SD for localization rather than generation. Furthermore, our method can still leverage diffusion models to generate images, which may lead to improved performance.

\begin{table*}[!tbp]
\centering
\caption{Synthetic dataset performance on ImageNet-1K with various teacher-student architectures under IPC=50. Our proposed method shows significant improvement with both CNNs and ViTs. ``\underline{  }\underline{ }\underline{ }'' denotes the best performance of student network.}
\scalebox{0.85}{
\begin{tabular}{lccccccc}
\toprule
\multirow{2}{*}{Teacher Network} & \multicolumn{7}{c}{Student Network} \\
\cmidrule{2-8}
 & ResNet-18 & ResNet-50  & ResNet-101 & MobileNet-V2 & EfficientNet-B0 & Swin-T & ViT-B \\
\midrule
ResNet-18 & \underline{61.0} & 63.1 & 65.7 & 53.6 & 60.4 & \underline{61.8} & \underline{63.7} \\
ResNet-50 & 52.2 & \underline{64.8} & \underline{66.7} & 48.9 & 56.2 & 51.3 & 60.1 \\
ResNet-101 & 50.1 & 63.7 & 66.5 & 46.1 & 55.5 & 42.8 & 50.5 \\
MobileNet-V2 & 55.2 & 62.5 & 65.0 & \underline{55.3} & \underline{60.5} & 57.2 & 62.3 \\
EfficientNet-B0 & 45.8 & 59.1 & 61.1 & 45.0 & 58.4 & 51.7 & 60.3 \\
Swin-T & 34.4 & 51.2 & 52.7 & 35.1 & 44.1 & 51.1 & 50.2 \\
ViT-B & 31.2 & 46.9 & 49.3 & 32.7 & 40.0  & 41.3 & 46.2\\
\bottomrule
\end{tabular}
}
\label{tab:cross}
\end{table*}

\subsection{Cross-architecture Generalization}
Following previous work, we evaluate the performance of the synthetic dataset using different teacher-student combinations. 
It can be observed that, when using similar network architectures as teacher-student pairs (e.g., both CNNs), shallower networks generally yield better performance as teacher models, while the larger networks show stronger learning ability and obtain the best performance when the size of synthetic datasets increases.

We further compare the performance of the synthetic dataset on extremely different network architectures, e.g., CNNs and ViTs. As a student network, the ViT-based networks leverages its global attention mechanism to attain the comparable performance with more realistic images. However, as a teacher network, ViT-based networks does not have strong ability to teach the student networks, yielding weak performance. 
Additionally, due to the low-pass filter characteristics of ViTs, effective learning becomes challenging when the sample size is limited and class-specific features are incomplete.

\begin{figure*}[!t]
  \centering
  \includegraphics[width=\textwidth]{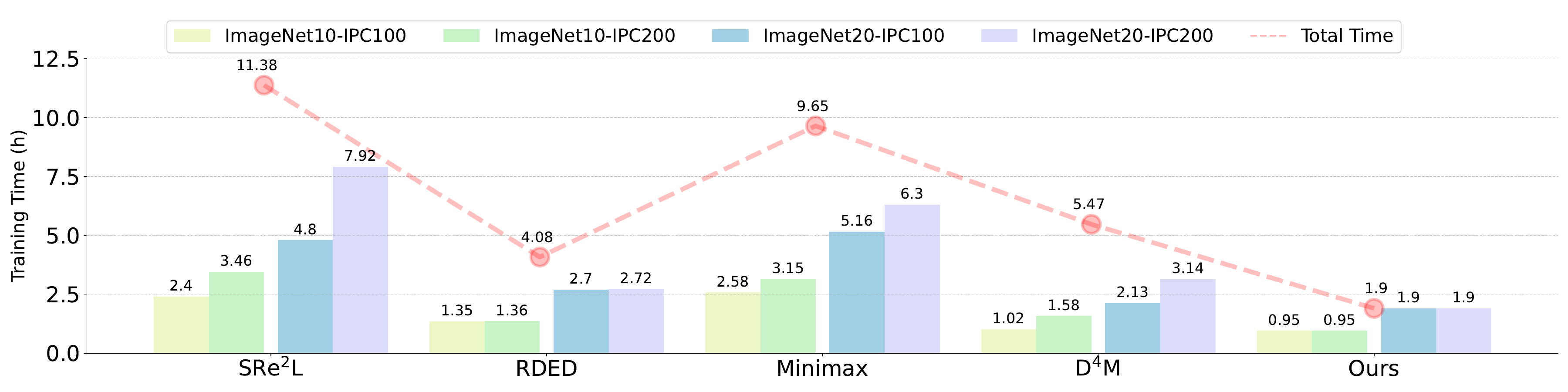}
  \caption{The separate time cost of individual tasks and total time cost of finishing all the tasks. Our method achieves a constant time cost due to one-step distillation process, which eliminates the need for repeated distillation.}
  \label{fig:efficiency}
\end{figure*}

\begin{table}[!tbp]
    \centering
    \caption{
        We present an analysis of the generalization capabilities of various distillation methods under different settings, where \Checkmark denote ``Satisfactory''.
    }
    \scalebox{1.1}{
        \begin{tabular}{l|cccccccc}
            \toprule                             
               &  DATM    &  SRe$^2$L     & Minimax  & IGD  & D$^4$M & MGD$^3$  & RDED  & Ours \\ \midrule
            Large scale & - & \Checkmark & \Checkmark & \Checkmark & \Checkmark & \Checkmark & \Checkmark & \Checkmark\\
            Zero shot & -   & - & - & -  & \Checkmark & \Checkmark & - & \Checkmark   \\
            Class extension  & -  & -   & - & -  & \Checkmark  & \Checkmark & \Checkmark & \Checkmark\\
            IPC extension & -   & \Checkmark & \Checkmark  & \Checkmark   & - & -  & \Checkmark & \Checkmark   \\ 
            \bottomrule
        \end{tabular}
    }
    \label{tab:limitation}
\end{table}

\subsection{Training Efficiency}
Distinct from all previous methods, our proposed method achieves a fully unrestricted distillation process and demonstrates SOTA performance. The limitations of existing dataset distillation methods are outlined as follows, and \Cref{tab:limitation} represents a comprehensive comparison: 
\begin{itemize} 
\item {\bf Large scale. }information-matching dataset distillation methods are impractical for large-scale datasets due to massive time cost.
\item {\bf Zero shot. }Relying on pre-trained classifiers require selecting an appropriate model architecture and training it for the target dataset. Minimax is also constrained by DiT pre-trained on ImageNet-1K. 
\item {\bf Class extension. }Repeated distillation processes are required when dealing with different class-combination. For example, in ImageNet-1K, a 10-class classification task can result in $C_{1000}^{10}$ possible combinations. 
\item {\bf IPC extension. }Varying IPC settings necessitate multiple distillation. While D$^4$M addresses the class-combination issue, it necessitates recalculating different prototypes with different IPC settings.
\end{itemize}

For a more comprehensive comparison, we evaluate the total synthesis time of different methods under challenging compression settings. As shown in \Cref{fig:efficiency}, our one-step distillation method achieves significant efficiency across various tasks, demonstrating a significant improvement in computational efficiency compared to previous approaches.

\subsection{Ablation Study}

\textbf{Clustering Strategy. }
To investigate the effectiveness of the two main stages of our proposed method, we conduct an ablation study to investigate whether clustering operations enhance the representative of synthetic datasets. Moreover, we conduct experiments to explore the differences between using CLIP features and DIFT during the clustering process. The experimental results as shown in \Cref{tab:cluster}, demonstrates that the performance of the synthetic dataset improves further after incorporating the clustering operation. This indicates that reducing sample randomness enables the model to learn more representative information. Additionally, we found DIFT more effective for achieving clustering compared to CLIP feature. 

\begin{table}[!t]
\centering
\caption{Performance Comparison on different data selecting strategys. Cluster-CLIP and Cluster-DIFT represent clustering with CLIP feature and DIFT respectively.}
\scalebox{1.}{
\begin{tabular}{lccc}
\toprule
Ablation & IPC-10 & IPC-50 & IPC-100 \\
\midrule
 & \multicolumn{3}{c}{Dataset: ImageWoof} \\
 \cmidrule{2-4}
w/o Cluster & 41.8 & 70.4 & 79.1 \\
w/ Cluster-CLIP & 39.8 & 69.5 & 79.5 \\
w/ Cluster-DIFT & \textbf{44.5(+2.7)} & \textbf{73.7(+3.3)} & \textbf{80.1(+1.0)} \\
\midrule
 & \multicolumn{3}{c}{Dataset: ImageNet-1K} \\
 \cmidrule{2-4}
w/o Cluster & 43.6 & 58.1 & 62.0 \\
w/ Cluster-CLIP & 42.1 & 59.0 & 62.7 \\
w/ Cluster-DIFT & \textbf{46.1(+2.5)} & \textbf{61.0(+1.9)} & \textbf{63.4(+1.4)} \\
\bottomrule
\end{tabular}}
\label{tab:cluster}
\vspace{-1em}
\end{table}

\begin{wraptable}{r}{0.45\linewidth}
    \centering
    \vspace{-1em}
    \caption{Ablation study of the range of time steps on ImageNet-1K under IPC=50. }
    \label{tab:time_range}
    \resizebox{\linewidth}{!}{
    \begin{tabular}{cccccc}
\toprule
 & \multicolumn{5}{c}{End} \\
 \cmidrule{2-6}
 Start & 0.1 & 0.3 & 0.5 & 0.7 & 0.9 \\
\midrule
0.1 & - & 58.9 & 59.3 & \textbf{61.0} & 59.7\\
0.3 & - & - & 58.8 & 58.9 & 59.6 \\
0.5 & - & - & 58.1 & 58.6 & 59.0 \\
0.7 & - & - & - & - & 57.9 \\
\bottomrule
\end{tabular}}
    \vspace{-1.0em}
\end{wraptable}
\textbf{Diffusion Time Intervals. }
To calculate the difference in predicted loss corresponding to the UNet when using different prompts (i.e., with and without the label text), we sample multiple diffusion time steps \(t\) as time conditions and average the results. 
As shown in \cref{tab:time_range}, we selected different minimum and maximum boundaries for the time step range and evaluated the corresponding performance of the synthetic dataset. It can be observed that the dataset achieves the best performance when intermediate time steps are employed, while the inclusion of small time steps proves to be essential. This finding aligns with conclusions drawn in recent studies \cite{li2023your}\cite{tang2023emergent}, which highlight that early diffusion steps capture fine-grained details crucial for downstream tasks, whereas overly large time steps may introduce excessive noise that degrades representation quality. In practical applications, considering the generality of ImageNet-1K, we applied the same optimized range of diffusion time steps to all of its subsets (e.g., ImageWoof). 

\begin{figure}[!t]
  \centering
  \includegraphics[width=\linewidth]{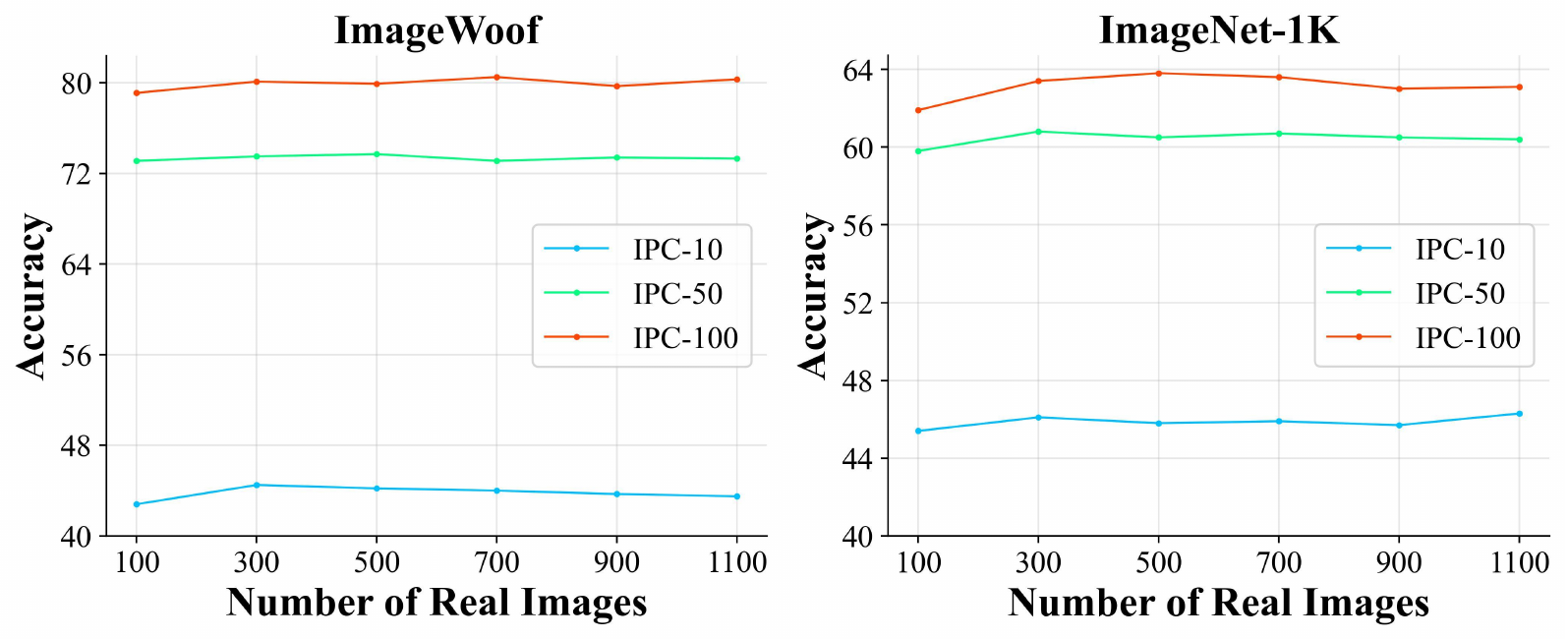}
  \caption{Ablation study on the number of real images used.}
  \label{fig:ablation-number}
  \vspace{-1.2em}
\end{figure}

\textbf{Number of Real Images. }
Although performing the distillation operation once on the entire dataset offers the potential for one-step processing, the substantial time overhead makes this approach difficult to apply in practice.
To address this issue, we explore the effect of the number of original images used for key patch extraction on performance.
As shown in \Cref{fig:ablation-number}, we randomly select images as a subset of the original dataset and evaluate the corresponding performance. Given that random selection is an unbiased data sampling method, our proposed achieves performance comparable to others, even when the number of original samples is limited. When 300 images are randomly selected from original dataset as input, the performance is nearly on par with the full dataset, resulting in a $4\times$ speedup.

\begin{wrapfigure}{r}{0.5\linewidth}
    \centering
    \includegraphics[width=1\linewidth]{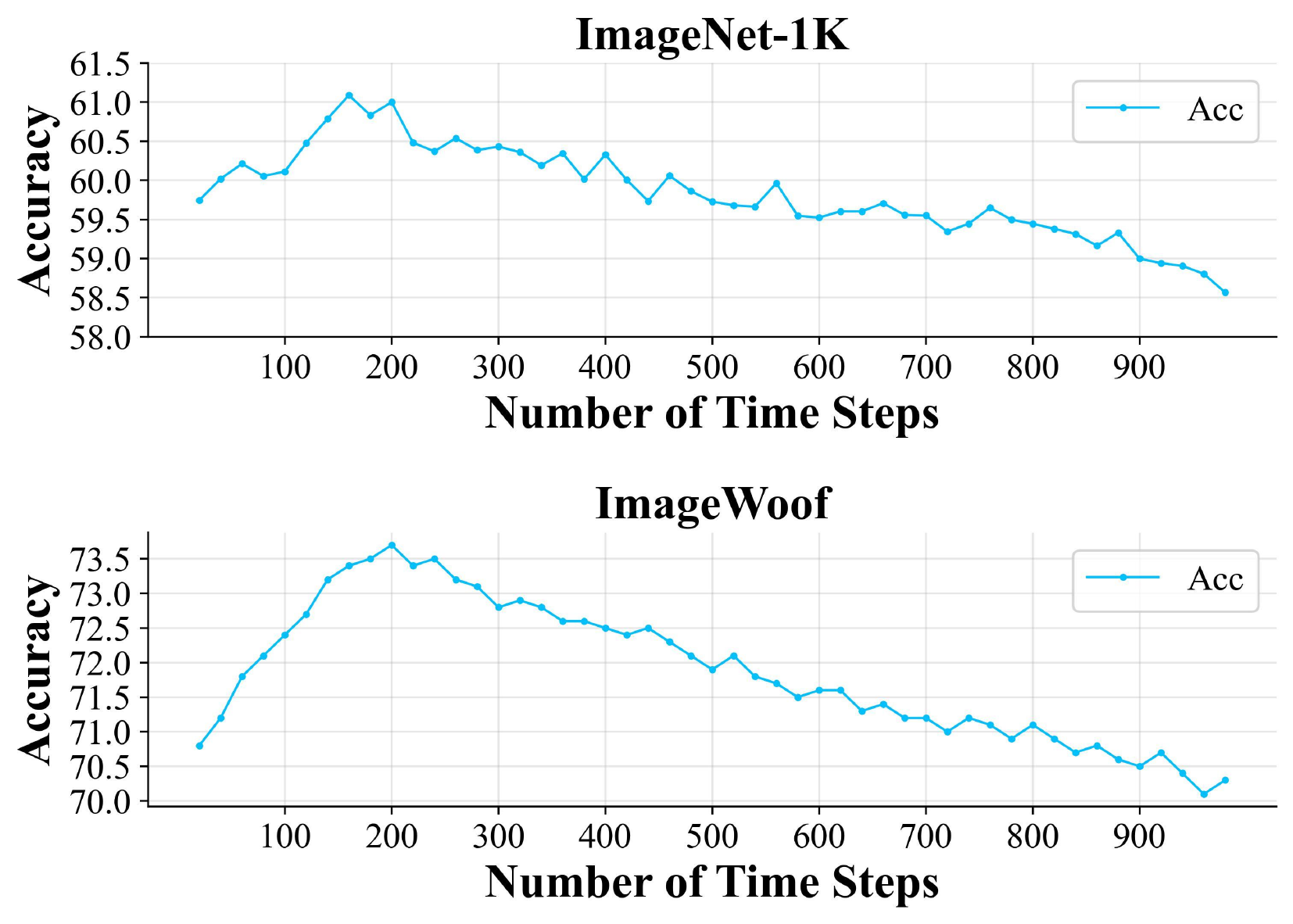}
    \caption{Ablation study of time steps used to calculate the predicted noise under IPC=50.}
  \label{fig:ablation-dift}
    \vspace{-2em}
\end{wrapfigure}
\textbf{Time Steps on Computation of DIFT. }
An important component of our proposed framework involves calculating the DIFT \cite{tang2023emergent} for all patches within the same class and using it as the feature for clustering. The process of calculating DIFT first requires performing a diffusion operation on the input patch at a specific time step. We conduct ablation experiments to investigate the impact of using different time steps. 
The experimental results are shown in \cref{fig:ablation-dift}.  We hypothesis the phenomenon is because the earlier time steps of the diffusion model focus more on the low-frequency features of the image, which often provide more effective information for classification tasks. In practical applications, we choose $t=160$ as the general setting.

%% file: sec/6_conclusion.tex
\section{Conclusion}
We propose a novel approach that operates orthogonally to existing diffusion-based generative dataset distillation methods, addressing the challenge of distribution mismatch between pre-trained diffusion models and target datasets. Through a two-stage framework, our method achieves a highly efficient, unrestricted one-step distillation process, eliminating the need for repeated distillations under varied settings. Extensive experiments demonstrate that our method significantly improves performance on large-scale datasets and deeper models, while achieving competitive results on low-resolution datasets. Additionally, our work provides a fresh perspective on leveraging diffusion models for dataset distillation.
\paragraph{\textbf{Acknowlegement.} }This work is supported in part by the National Science and Technology Major Project under grant 2025ZD1601000, National Natural Science Foundation of China under grant 62301189, 62576122,62571298, Guangdong Basic and Applied Basic Research Foundation under grant 2026A1515011139.


%% file: sec/X_suppl.tex
\setcounter{section}{0}
\renewcommand{\thesection}{\Alph{section}}

\section{More Experimental Results}

\subsection{Hard Label Performance}
To further investigate the quality of synthetic datasets generated by different distillation methods, we conduct an experiment with one-hot labels. 

As shown in \cref{tab:hard}, SRe$^2$L exhibit extreme poor performance, Due to the lack of diversity of generated images caused by using only cross-entropy loss. For Minimax and RDED, the synthetic datasets also exhibit concentration of class centers which means the loss of diversity, leading a performance degradation especially on large IPC settings. Despite D$^4$M successfully enhance diversity of synthetic datasets by using a SD trained on LAION, the misalignment phenomenon observed between images and corresponding class labels, leading to a significant performance drop when using hard label. In contrast, our method achieve a balance between representative and diversity which has also been verified using t-SNE visualization, thus a strong performance achieved using only hard label.

\begin{table}[!htbp]
\centering
\caption{Hard label performance on ImageNet-1K with different methods. }
\scalebox{1.1}{
\begin{tabular}{lcccccc}
\toprule
 IPC & Random & SRe$^2$L & Minimax & D$^4$M & RDED & Ours \\ \midrule
 1  & 1.1$\pm$0.3 & 0.2 $\pm$0.1   & 1.3 $\pm$ 0.7 & 0.7 $\pm$ 0.4  &  1.1 $\pm$ 0.3 & \textbf{1.7 $\pm$ 0.3}\\
10 & 7.3 $\pm$0.2 & 1.4 $\pm$ 0.2  & 8.2 $\pm$ 0.4 &  4.6 $\pm$ 0.5 & \textbf{11.4 $\pm$ 0.4} & 10.5 $\pm$ 0.3  \\
50 & 30.9 $\pm$0.1 & 3.2 $\pm$ 0.3  & 31.3 $\pm$ 0.4 & 17.6 $\pm$ 0.2 & 30.1 $\pm$ 0.3 & \textbf{33.2 $\pm$ 0.1} \\
100 & 41.0 $\pm$0.3 &  4.1 $\pm$ 0.3 & 40.5 $\pm$ 0.2  & 27.3 $\pm$ 0.5  & 38.4 $\pm$ 0.2 & \textbf{42.1 $\pm$ 0.2}  \\ 
\bottomrule
\end{tabular}
}
\label{tab:hard}
\end{table}

\begin{table*}[!tbp]
\centering
\caption{Synthetic dataset performance on ImageNet-1K with various teacher-student architectures under IPC=10. }
\scalebox{0.8}{
\begin{tabular}{lccccccc}
\toprule
\multirow{2}{*}{Teacher Network} & \multicolumn{7}{c}{Student Network} \\
\cmidrule{2-8}
 & ResNet-18 & ResNet-50  & ResNet-101 & MobileNet-V2 & EfficientNet-B0 & Swin-T & ViT-B \\
\midrule
ResNet-18 & \underline{46.1} & \underline{50.1} & \underline{51.2} & 36.1 & \underline{46.1} & \underline{31.2} & \underline{22.3} \\
ResNet-50 & 34.6 & 44.0 & 42.2 & 27.8 & 38.1 & 27.4 & 19.8 \\
ResNet-101 & 33.2 & 41.4 & 42.5 & 28.2 & 37.0 & 26.1 & 17.0 \\
MobileNet-V2 & 40.1 & 48.3 & 46.7 & \underline{37.5} & 46.0 & 30.1 & 20.5 \\
EfficientNet-B0 & 29.8 & 38.4 & 38.7 & 27.4 & 38.9 & 30.3 & 21.3 \\
Swin-T & 18.9 & 30.2 & 31.4 & 21.6 & 31.7 & 26.7 & 16.0 \\
ViT-B & 16.8 & 27.8 & 28.9 & 19.2 & 28.9  & 23.6 & 15.0 \\
\bottomrule
\end{tabular}}
\label{tab:cross_10}
\end{table*}

\subsection{Generalization Ability}
To better investigate the generalization capability of our proposed method under different settings, we evaluated its performance using various teacher-student model pairs at IPC=10. Unlike the IPC=50 setting, each image generated is composed of four patches. 

As shown in \cref{tab:cross_10}, our proposed method still demonstrates strong performance on CNNs. Additionally, it can be observed that ResNet-18 is the most suitable architecture to be used as a teacher model when the sample size is smaller. However, for ViTs, due to their low-pass filter characteristics, effective learning becomes challenging when the sample size is limited and class-specific features are incomplete. The best performance achieved when using DiT as a student model is only 50\% of that of the CNN student model under the same settings.

\section{Additional Ablation Studies}

\begin{figure}[!t]
  \centering
  \includegraphics[width=\linewidth]{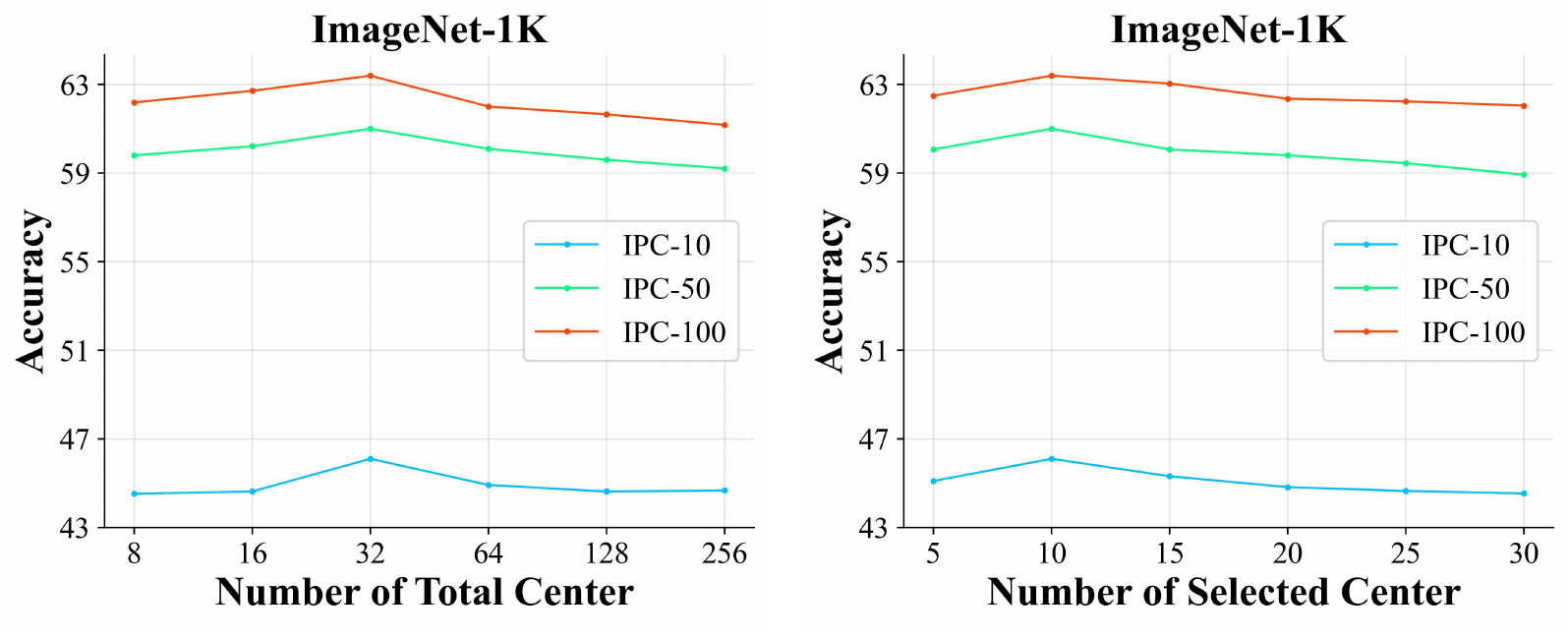}
  \caption{Ablation study of the number of cluster center M and the number of selected cluster N on ImageNet-1K.}
  \label{fig:ablation-cluster}
\end{figure}

\subsection{Number of Cluster Centers}
To enhance the representativeness of the synthetic dataset, we performed clustering on all patches to extract the most class-relevant visual elements. To gain deeper insights into the impact of the clustering operation on the performance of the synthetic dataset, we conduct additional ablation experiments focusing on the number of cluster centers. Specifically, we evaluate the performance of the synthetic dataset using different numbers of cluster centers $M$, as well as the number $N$ of top clusters used for extracting patches after clustering. In practical applications, when IPC $\leq$ 10, the number of patches provided by each cluster is 4*IPC // $N$, otherwise the number is IPC // N. The remaining patches are selected sequentially from the unused ones based on $R(p|c)$ of each patch $p$ in descending order. 

The experimental results are shown in \cref{fig:ablation-cluster}, indicating that when $N$ is fixed, both too few or too many clusters lead to a failure in balancing diversity and representativeness in the synthetic dataset, resulting in a decline in performance. A similar trend was observed in the ablation experiment where $M$ was fixed and $N$ varied.

\begin{wrapfigure}{r}{0.5\linewidth}
    \centering
    \vspace{-4em}
    \includegraphics[width=1\linewidth]{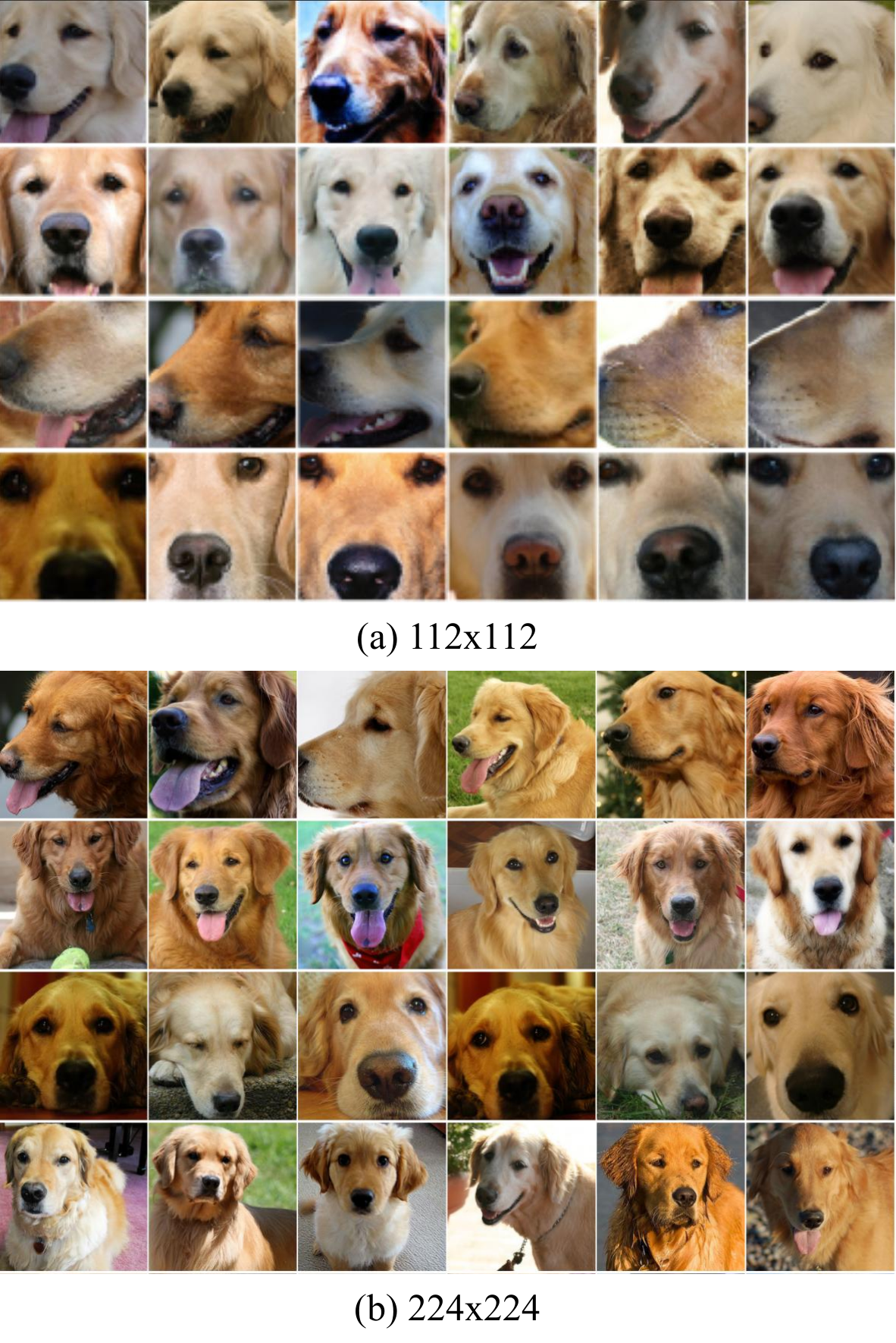}
   \caption{Visualizations of the patches with different cropped size within class ``Golden Retriever''.}
  \label{fig:comp-patch}
    \vspace{-2em}
\end{wrapfigure}
\subsection{Impact of Resizing}
When using diffusion to process images, the image size often has a certain degree of impact. On the other hand, when calculating $R(p|c)$ for each patch, we used a fixed-size average pooling kernel to obtain the mean value of the patch. Therefore, we designed an ablation experiment to explore whether different input image sizes and patch sizes significantly affect the performance of the synthetic dataset. In practical applications, we proportionally scale the original image such that its size of shortest edge matches the selected size. For different patch sizes, under the setting of IPC=10, we scale all patches to $112 \times 112$ in order to concatenate 4 patches into a $224 \times 224$ image. Under the setting of IPC=50, we scale all patches to $224 \times 224$. 

The experimental results, as shown in \cref{tab:abl_size}, indicate that when the original image size closely matches the standard output size of SD v1.5, better performance are achieved. Additionally, when the size of the cropped patches is closer to the scaled original image size, the image tends to exhibit more complete class-relevant features, leading to better performance. Consider the balance between performance and efficiency, We choose 256x256 as the size of the real images, and $224 \times 224$ as the size of the patches. Comparison of visualization with different crop sizes when using 256x256 as the size of the real images are shown in \cref{fig:comp-patch}, An excessive size difference between patches and the original image can cause the generated dataset to overly focus on low-frequency features (e.g., noses of dogs), leading to performance degradation.

\begin{table}[!ht]
\centering
\caption{Ablation study of different size of real images and patches on ImageWoof}
\scalebox{1.2}{
\begin{tabular}{lccc}
\toprule
Patch Size  & IPC-10 & IPC-50 & IPC-100 \\
\midrule
 & \multicolumn{3}{c}{Image Size: $128 \times 128$} \\
 \cmidrule{2-4}
$112 \times 112$ & 42.6 & 70.2 & 77.8 \\
\midrule
 & \multicolumn{3}{c}{Image Size: $256 \times 256$} \\
 \cmidrule{2-4}
$112 \times 112$ & 40.4 & 67.4 & 76.8 \\
$224 \times 224$ & 44.5 & 73.7 & 80.1 \\
\midrule
 & \multicolumn{3}{c}{Image Size: $512 \times 512$} \\
 \cmidrule{2-4}
$112 \times 112$ & 38.6 & 65.3 & 71.4 \\
$224 \times 224$ & 43.1 & 69.5 & 78.4 \\
$448 \times 448$ & 45.1 & 73.2 & 81.0 \\
\bottomrule
\end{tabular}}
\label{tab:abl_size}
\end{table}

\section{Visualizations}
\begin{figure*}[!t]
  \centering
  \includegraphics[width=\textwidth]{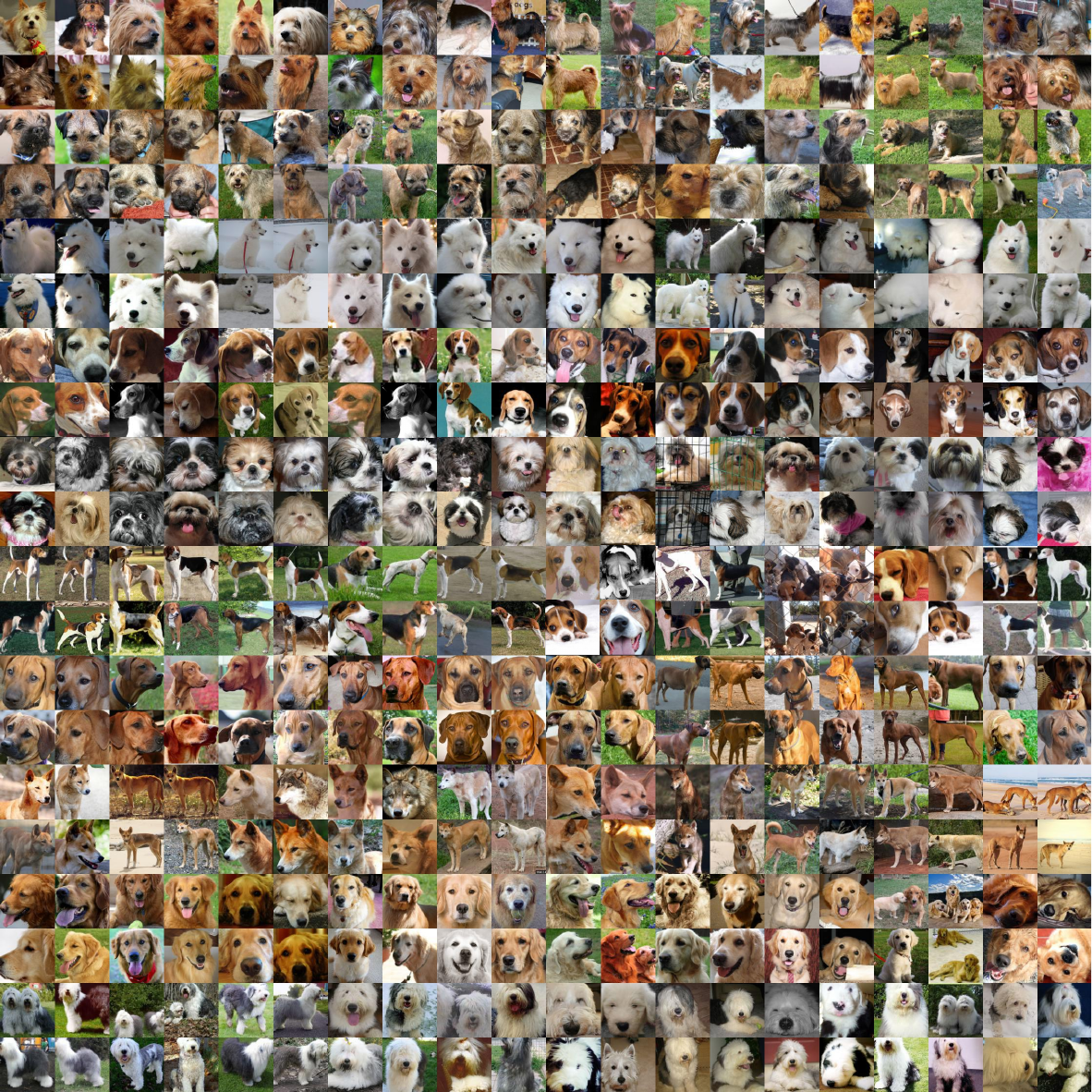}
  \caption{Visualization of ImageWoof under IPC=10.}
  \label{fig:vis-woof-10}
\end{figure*}

\begin{figure*}[!t]
  \centering
  \includegraphics[width=\textwidth]{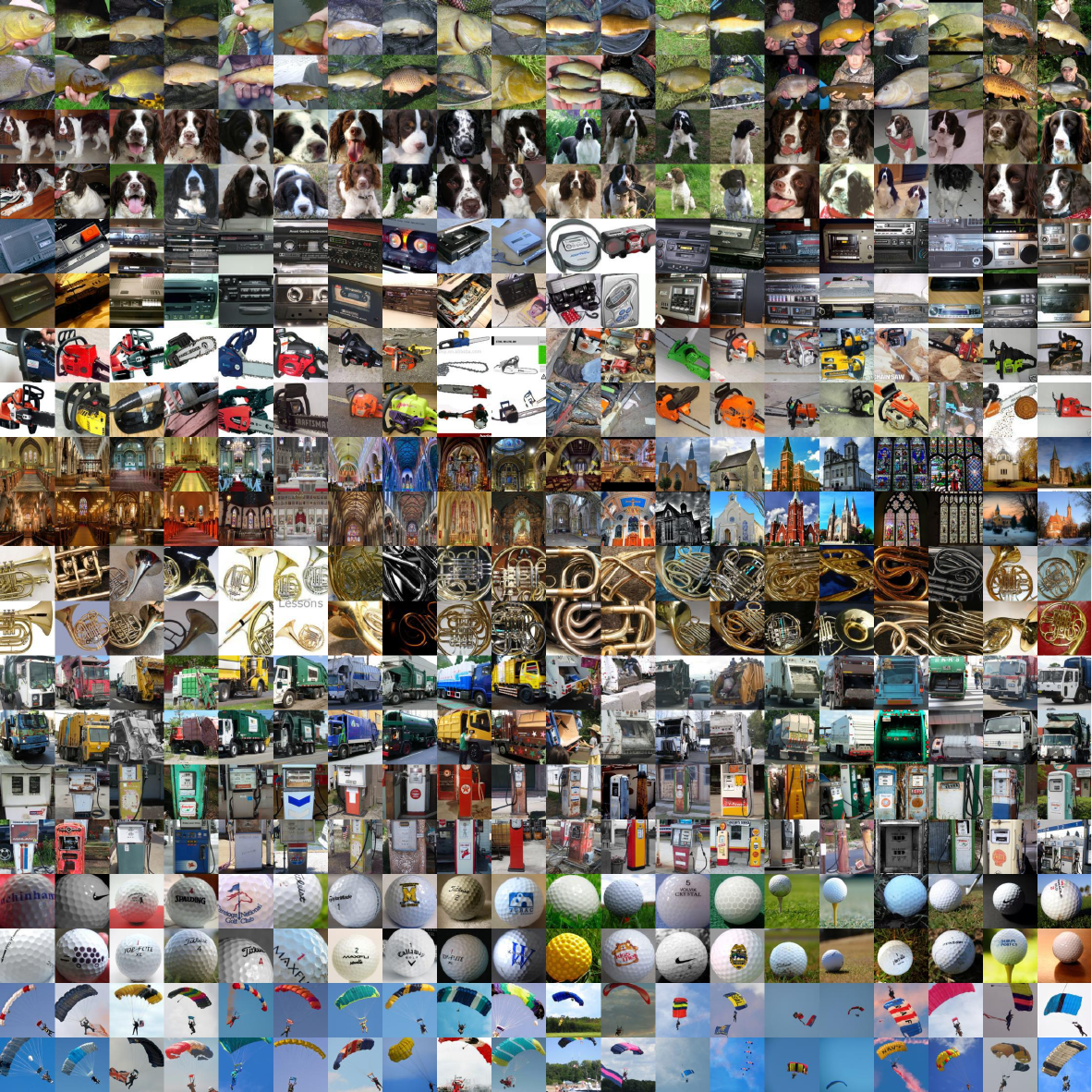}
  \caption{Visualization of ImageNette under IPC=10.}
  \label{fig:vis-nette-10}
\end{figure*}

\begin{figure*}[!t]
  \centering
  \includegraphics[width=\textwidth]{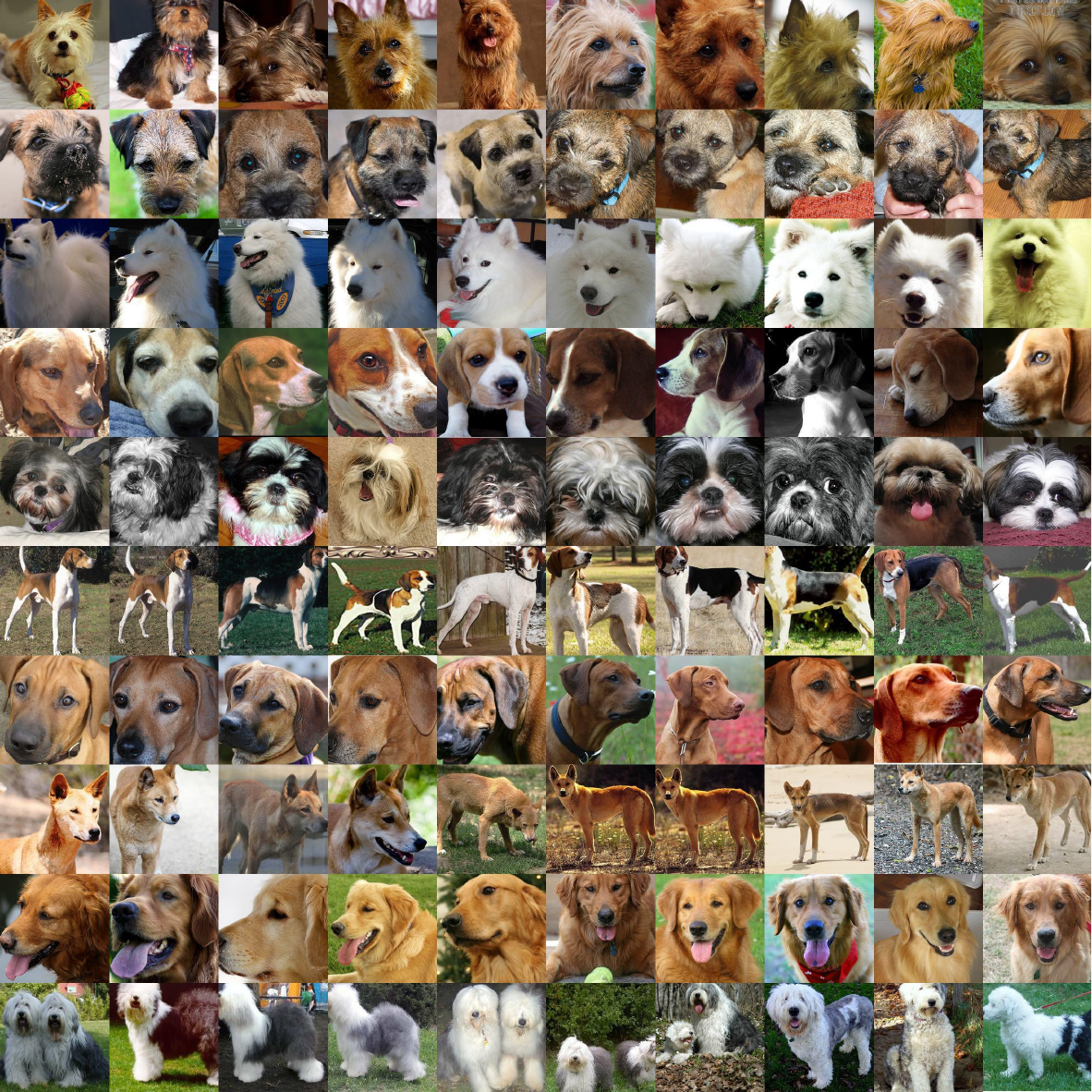}
  \caption{Partial visualization of ImageWoof under IPC=50.}
  \label{fig:vis-woof-50}
\end{figure*}

\begin{figure*}[!t]
  \centering
  \includegraphics[width=\textwidth]{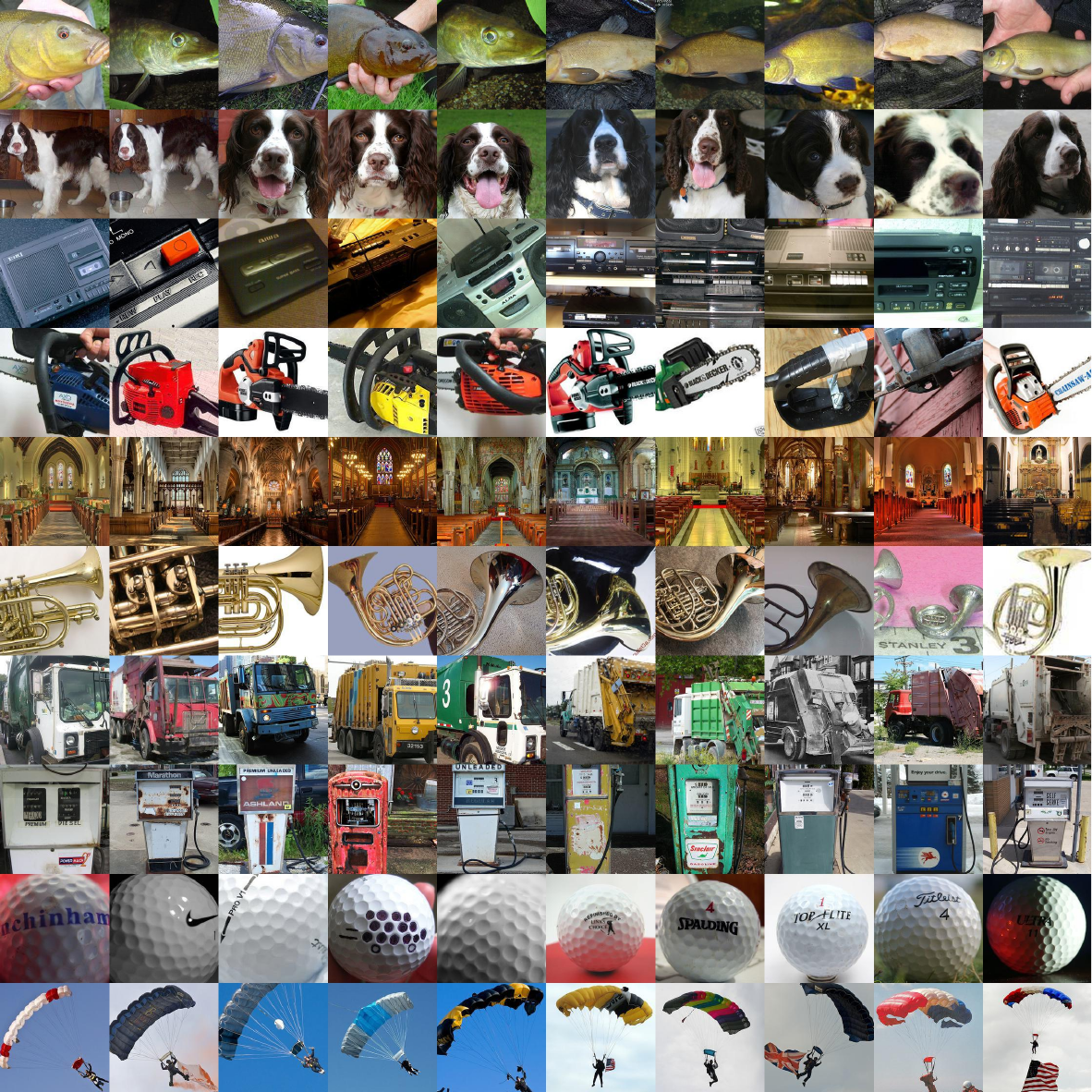}
  \caption{Partial visualization of ImageNette under IPC=50.}
  \label{fig:vis-nette-50}
\end{figure*}

\begin{figure*}[!t]
  \centering
  \includegraphics[width=\textwidth]{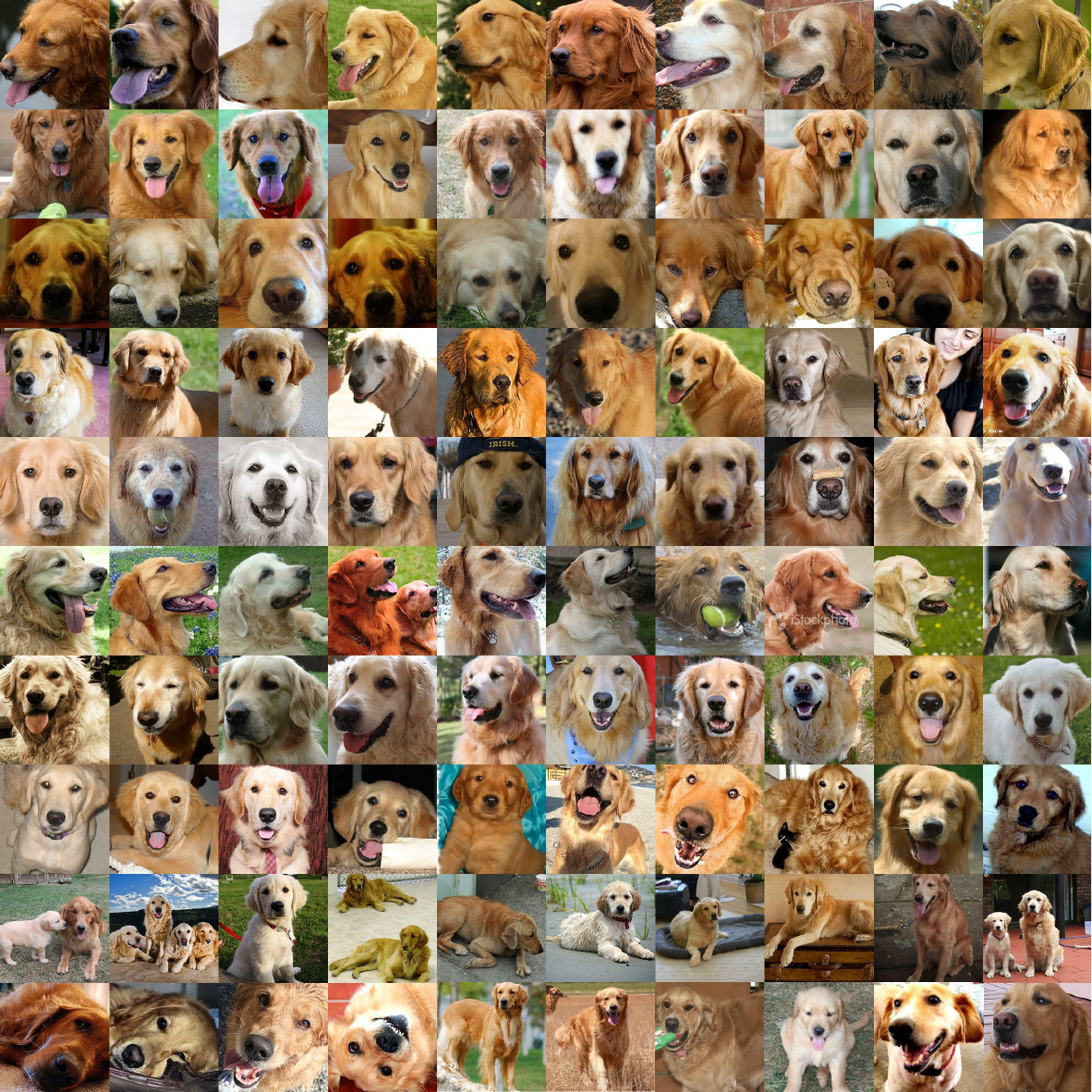}
  \caption{Visualization of class ``Golden Retriever'' under IPC=100.}
  \label{fig:vis-woof-100}
\end{figure*}

To facilitate a more intuitive understanding of our methodology, we present the complete synthetic datasets for ImageWoof and ImageNette at IPC=10 as illustrated in \cref{fig:vis-woof-10,fig:vis-nette-10}, as well as the partial synthetic dataset at IPC=50, depicted in \cref{fig:vis-woof-50,fig:vis-nette-50}. Furthermore, we have visualized all images belonging to the ``Golden Retriever" class at IPC=100, as shown in \cref{fig:vis-woof-100}.


\clearpage

\clearpage

\clearpage

\clearpage

\clearpage

\clearpage


%% file: main.bib
@String(ICLR = {Int. Conf. Learn. Represent.})

@String(ICLR  = {ICLR})

@article{lei2023comprehensive,
  title={A comprehensive survey to dataset distillation},
  author={Lei, Shiye and Tao, Dacheng},
  journal={arXiv preprint arXiv:2301.05603},
  year={2023}
}

@inproceedings{chen2025influence,
  title={Influence-guided diffusion for dataset distillation},
  author={Chen, Mingyang and Du, Jiawei and Huang, Bo and Wang, Yi and Zhang, Xiaobo and Wang, Wei},
  booktitle={The Thirteenth International Conference on Learning Representations},
  year={2025}
}

@article{li2025dd,
  title={Dd-ranking: Rethinking the evaluation of dataset distillation},
  author={Li, Zekai and Zhong, Xinhao and Khaki, Samir and Liang, Zhiyuan and Zhou, Yuhao and Shi, Mingjia and Wang, Ziqiao and Zhao, Xuanlei and Zhao, Wangbo and Qin, Ziheng and others},
  journal={arXiv preprint arXiv:2505.13300},
  year={2025}
}

@inproceedings{zhong2025going,
  title={Going beyond feature similarity: effective dataset distillation based on class-aware conditional mutual information},
  author={Zhong, Xinhao and Chen, Bin and Fang, Hao and Gu, Xulin and Xia, Shu-Tao and Yang, En-Hui},
  booktitle={International Conference on Learning Representations},
  volume={2025},
  pages={36598--36615},
  year={2025}
}

@article{gu2025temporal,
  title={Temporal saliency-guided distillation: A scalable framework for distilling video datasets},
  author={Gu, Xulin and Zhong, Xinhao and Wei, Zhixing and Zhou, Yimin and Sun, Shuoyang and Chen, Bin and Wang, Hongpeng and Luo, Yuan},
  journal={arXiv preprint arXiv:2505.20694},
  year={2025}
}

@article{zhong2025rectified,
  title={Rectified Decoupled Dataset Distillation: A Closer Look for Fair and Comprehensive Evaluation},
  author={Zhong, Xinhao and Sun, Shuoyang and Gu, Xulin and Zhu, Chenyang and Chen, Bin and Wang, Yaowei},
  journal={arXiv preprint arXiv:2509.19743},
  year={2025}
}

@article{zhao2025taming,
  title={Taming Diffusion for Dataset Distillation with High Representativeness},
  author={Zhao, Lin and Wu, Yushu and Jiang, Xinru and Gu, Jianyang and Wang, Yanzhi and Xu, Xiaolin and Zhao, Pu and Lin, Xue},
  journal={arXiv preprint arXiv:2505.18399},
  year={2025}
}

@article{chan2025mgd,
  title={MGD3: Mode-Guided Dataset Distillation using Diffusion Models},
  author={Chan-Santiago, Jeffrey A and Tirupattur, Praveen and Nayak, Gaurav Kumar and Liu, Gaowen and Shah, Mubarak},
  journal={arXiv preprint arXiv:2505.18963},
  year={2025}
}

@article{shao2024elucidating,
  title={Elucidating the design space of dataset condensation},
  author={Shao, Shitong and Zhou, Zikai and Chen, Huanran and Shen, Zhiqiang},
  journal={Advances in neural information processing systems},
  volume={37},
  pages={99161--99201},
  year={2024}
}

@inproceedings{siglidis2024diffusion,
  title={Diffusion Models as Data Mining Tools},
  author={Siglidis, Ioannis and Holynski, Aleksander and Efros, Alexei A and Aubry, Mathieu and Ginosar, Shiry},
  booktitle={European Conference on Computer Vision},
  pages={393--409},
  year={2024},
  organization={Springer}
}

@article{dosovitskiy2020image,
  title={An image is worth 16x16 words: Transformers for image recognition at scale},
  author={Dosovitskiy, Alexey and Beyer, Lucas and Kolesnikov, Alexander and Weissenborn, Dirk and Zhai, Xiaohua and Unterthiner, Thomas and Dehghani, Mostafa and Minderer, Matthias and Heigold, Georg and Gelly, Sylvain and others},
  journal={arXiv preprint arXiv:2010.11929},
  year={2020}
}

@inproceedings{sun2024diversity,
  title={On the diversity and realism of distilled dataset: An efficient dataset distillation paradigm},
  author={Sun, Peng and Shi, Bei and Yu, Daiwei and Lin, Tao},
  booktitle={Proceedings of the IEEE/CVF Conference on Computer Vision and Pattern Recognition},
  pages={9390--9399},
  year={2024}
}

@inproceedings{gu2024efficient,
  title={Efficient dataset distillation via minimax diffusion},
  author={Gu, Jianyang and Vahidian, Saeed and Kungurtsev, Vyacheslav and Wang, Haonan and Jiang, Wei and You, Yang and Chen, Yiran},
  booktitle={Proceedings of the IEEE/CVF Conference on Computer Vision and Pattern Recognition},
  pages={15793--15803},
  year={2024}
}

@inproceedings{su2024d,
  title={D\^{} 4: Dataset Distillation via Disentangled Diffusion Model},
  author={Su, Duo and Hou, Junjie and Gao, Weizhi and Tian, Yingjie and Tang, Bowen},
  booktitle={Proceedings of the IEEE/CVF Conference on Computer Vision and Pattern Recognition},
  pages={5809--5818},
  year={2024}
}

@article{zhong2024hierarchical,
  title={Hierarchical Features Matter: A Deep Exploration of GAN Priors for Improved Dataset Distillation},
  author={Zhong, Xinhao and Fang, Hao and Chen, Bin and Gu, Xulin and Dai, Tao and Qiu, Meikang and Xia, Shu-Tao},
  journal={arXiv preprint arXiv:2406.05704},
  year={2024}
}

@article{he2024multisize,
  title={Multisize dataset condensation},
  author={He, Yang and Xiao, Lingao and Zhou, Joey Tianyi and Tsang, Ivor},
  journal={arXiv preprint arXiv:2403.06075},
  year={2024}
}

@inproceedings{kim2022dataset,
  title={Dataset condensation via efficient synthetic-data parameterization},
  author={Kim, Jang-Hyun and Kim, Jinuk and Oh, Seong Joon and Yun, Sangdoo and Song, Hwanjun and Jeong, Joonhyun and Ha, Jung-Woo and Song, Hyun Oh},
  booktitle={International Conference on Machine Learning},
  pages={11102--11118},
  year={2022},
  organization={PMLR}
}

@article{ho2020denoising,
  title={Denoising diffusion probabilistic models},
  author={Ho, Jonathan and Jain, Ajay and Abbeel, Pieter},
  journal={NeurIPS},
  year={2020}
}

@article{song2020denoising,
  title={Denoising diffusion implicit models},
  author={Song, Jiaming and Meng, Chenlin and Ermon, Stefano},
  journal={ICLR},
  year={2021}
}

@article{krizhevsky2009learning,
  title={Learning Multiple Layers of Features from Tiny Images},
  author={Krizhevsky, A},
  journal={Master's thesis, University of Tront},
  year={2009}
}

@inproceedings{li2023your,
  title={Your diffusion model is secretly a zero-shot classifier},
  author={Li, Alexander C and Prabhudesai, Mihir and Duggal, Shivam and Brown, Ellis and Pathak, Deepak},
  booktitle={Proceedings of the IEEE/CVF International Conference on Computer Vision},
  pages={2206--2217},
  year={2023}
}

@article{lloyd1982least,
  title={Least squares quantization in PCM},
  author={Lloyd, Stuart},
  journal={IEEE transactions on information theory},
  volume={28},
  number={2},
  pages={129--137},
  year={1982},
  publisher={IEEE}
}

@article{zhu2024instantswap,
  title={InstantSwap: Fast Customized Concept Swapping across Sharp Shape Differences},
  author={Zhu, Chenyang and Li, Kai and Ma, Yue and Tang, Longxiang and Fang, Chengyu and Chen, Chubin and Chen, Qifeng and Li, Xiu},
  journal={arXiv preprint arXiv:2412.01197},
  year={2024}
}

@article{zhu2024multibooth,
  title={MultiBooth: Towards Generating All Your Concepts in an Image from Text},
  author={Zhu, Chenyang and Li, Kai and Ma, Yue and He, Chunming and Xiu, Li},
  journal={arXiv preprint arXiv:2404.14239},
  year={2024}
}

@article{ho2022classifier,
  title={Classifier-free diffusion guidance},
  author={Ho, Jonathan and Salimans, Tim},
  journal={arXiv preprint arXiv:2207.12598},
  year={2022}
}

@inproceedings{ronneberger2015u,
  title={U-net: Convolutional networks for biomedical image segmentation},
  author={Ronneberger, Olaf and Fischer, Philipp and Brox, Thomas},
  booktitle={Medical image computing and computer-assisted intervention--MICCAI 2015: 18th international conference, Munich, Germany, October 5-9, 2015, proceedings, part III 18},
  pages={234--241},
  year={2015},
  organization={Springer}
}

@article{schuhmann2022laion,
  title={Laion-5b: An open large-scale dataset for training next generation image-text models},
  author={Schuhmann, Christoph and Beaumont, Romain and Vencu, Richard and Gordon, Cade and Wightman, Ross and Cherti, Mehdi and Coombes, Theo and Katta, Aarush and Mullis, Clayton and Wortsman, Mitchell and others},
  journal={Advances in Neural Information Processing Systems},
  volume={35},
  pages={25278--25294},
  year={2022}
}

@inproceedings{deng2009imagenet,
  title={Imagenet: A large-scale hierarchical image database},
  author={Deng, Jia and Dong, Wei and Socher, Richard and Li, Li-Jia and Li, Kai and Fei-Fei, Li},
  booktitle={2009 IEEE conference on computer vision and pattern recognition},
  pages={248--255},
  year={2009},
  organization={Ieee}
}

@article{tang2023emergent,
  title={Emergent correspondence from image diffusion},
  author={Tang, Luming and Jia, Menglin and Wang, Qianqian and Phoo, Cheng Perng and Hariharan, Bharath},
  journal={Advances in Neural Information Processing Systems},
  volume={36},
  pages={1363--1389},
  year={2023}
}

@inproceedings{rombach2022high,
  title={High-resolution image synthesis with latent diffusion models},
  author={Rombach, Robin and Blattmann, Andreas and Lorenz, Dominik and Esser, Patrick and Ommer, Bj{\"o}rn},
  booktitle={Proceedings of the IEEE/CVF conference on computer vision and pattern recognition},
  pages={10684--10695},
  year={2022}
}

@inproceedings{he2016deep,
  title={Deep residual learning for image recognition},
  author={He, Kaiming and Zhang, Xiangyu and Ren, Shaoqing and Sun, Jian},
  booktitle={Proceedings of the IEEE conference on computer vision and pattern recognition},
  pages={770--778},
  year={2016}
}

@article{zhao2021datasetdm,
  title={Dataset Condensation with Distribution Matching},
  author={Zhao, Bo and Bilen, Hakan},
  journal={arXiv preprint arXiv:2110.04181},
  year={2021}
}

@inproceedings{cazenavette2022dataset,
  title={Dataset distillation by matching training trajectories},
  author={Cazenavette, George and Wang, Tongzhou and Torralba, Antonio and Efros, Alexei A and Zhu, Jun-Yan},
  booktitle={Proceedings of the IEEE/CVF Conference on Computer Vision and Pattern Recognition},
  pages={4750--4759},
  year={2022}
}

@article{wang2018dataset,
  title={Dataset distillation},
  author={Wang, Tongzhou and Zhu, Jun-Yan and Torralba, Antonio and Efros, Alexei A},
  journal={arXiv preprint arXiv:1811.10959},
  year={2018}
}

@inproceedings{zhao2021dsa,
  title={Dataset condensation with differentiable siamese augmentation},
  author={Zhao, Bo and Bilen, Hakan},
  booktitle={International Conference on Machine Learning},
  pages={12674--12685},
  year={2021},
  organization={PMLR}
}

@article{cazenavette2023generalizing,
  title={Generalizing Dataset Distillation via Deep Generative Prior},
  author={Cazenavette, George and Wang, Tongzhou and Torralba, Antonio and Efros, Alexei A and Zhu, Jun-Yan},
  journal={arXiv preprint arXiv:2305.01649},
  year={2023}
}

@article{yin2024squeeze,
  title={Squeeze, recover and relabel: Dataset condensation at imagenet scale from a new perspective},
  author={Yin, Zeyuan and Xing, Eric and Shen, Zhiqiang},
  journal={Advances in Neural Information Processing Systems},
  volume={36},
  year={2024}
}

@article{he2024you,
  title={You only condense once: Two rules for pruning condensed datasets},
  author={He, Yang and Xiao, Lingao and Zhou, Joey Tianyi},
  journal={Advances in Neural Information Processing Systems},
  volume={36},
  year={2024}
}

@inproceedings{peebles2023scalable,
  title={Scalable diffusion models with transformers},
  author={Peebles, William and Xie, Saining},
  booktitle={Proceedings of the IEEE/CVF International Conference on Computer Vision},
  pages={4195--4205},
  year={2023}
}

@inproceedings{gidaris2018dynamic,
  title={Dynamic few-shot visual learning without forgetting},
  author={Gidaris, Spyros and Komodakis, Nikos},
  booktitle={Proceedings of the IEEE conference on computer vision and pattern recognition},
  pages={4367--4375},
  year={2018}
}

@article{le2015tiny,
  title={Tiny imagenet visual recognition challenge},
  author={Le, Ya and Yang, Xuan},
  journal={CS 231N},
  volume={7},
  number={7},
  pages={3},
  year={2015}
}
